\documentclass{article}

\usepackage{arxiv}

\usepackage[utf8]{inputenc} 
\usepackage[T1]{fontenc}    
\usepackage{hyperref}       
\usepackage{url}            
\usepackage{booktabs}       
\usepackage{amsfonts}       
\usepackage{nicefrac}       
\usepackage{microtype}      
\usepackage{lipsum}		
\usepackage{graphicx}
\usepackage{natbib}
\usepackage{doi}
\usepackage{amsmath} 
\usepackage{amssymb}  
\usepackage{multirow}
\usepackage{booktabs}
\usepackage[normalem]{ulem}
\usepackage[HTML, table]{xcolor}
\usepackage{nicematrix}
\usepackage{subcaption}

\usepackage{array}

\definecolor{myred}{HTML}{FFAAAA}
\definecolor{mylightred}{HTML}{FFDDDD}
\definecolor{mylightgreen}{HTML}{DDFFDD}
\definecolor{mygreen}{HTML}{AAFFAA}

\newcommand{\cc}[2]{%
  \cellcolor{#1}\textcolor{black}{#2}%
}

\title{HUydra: Full-Range Lung CT Synthesis via Multiple HU Interval Generative Modelling}

\author{ 
    \href{https://orcid.org/0009-0008-2879-7222}{\includegraphics[scale=0.06]{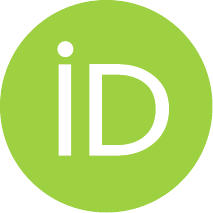}\hspace{1mm}Ant\'onio Cardoso}\\
    INESC TEC, Portugal\\
    Faculty of Engineering, University of Porto, Portugal\\
    Faculty of Sciences, University of Porto, Portugal\\
	\texttt{antonio.l.cardoso@inesctec.pt} \\
	\And
	\href{https://orcid.org/0009-0005-0332-2149}{\includegraphics[scale=0.06]{orcid.pdf}\hspace{1mm}Pedro Sousa}\\
    INESC TEC, Portugal\\
	\texttt{pedro.fernandes.sousa@inesctec.pt} \\
    \And
	\href{https://orcid.org/0000-0003-1681-2436}{\includegraphics[scale=0.06]{orcid.pdf}\hspace{1mm}Tania Pereira}\\
    INESC TEC, Portugal\\
    Faculty of Engineering, University of Porto,  Portugal\\
	\texttt{tania.pereira@inesctec.pt} \\
    \And
	\href{https://orcid.org/0000-0002-6193-8540}{\includegraphics[scale=0.06]{orcid.pdf}\hspace{1mm}H\'elder P. Oliveira}\\
    INESC TEC, Portugal\\
    Faculty of Sciences, University of Porto, Portugal\\
	\texttt{helder.f.oliveira@inesctec.pt} \\
}

\date{}

\begin{document}
\maketitle


\begin{abstract}

Currently, a central challenge and bottleneck in the deployment and validation of computer-aided diagnosis (CAD) models within the field of medical imaging is data scarcity. For lung cancer, one of the most prevalent types worldwide, limited datasets can delay diagnosis and have an impact on patient outcome. Generative AI offers a promising solution for this issue, but dealing with the complex distribution of full Hounsfield Unit (HU) range lung CT scans is challenging and remains as a highly computationally demanding task. This paper introduces a novel decomposition strategy that synthesizes CT images one HU interval at a time, rather than modelling the entire HU domain at once. This framework focuses on training generative architectures on individual tissue-focused HU windows, then merges their output into a full-range scan via a learned reconstruction network that effectively reverses the HU-windowing process. We further propose multi-head and multi-decoder models to better capture textures while preserving anatomical consistency, with a multi-head VQVAE achieving the best performance for the generative task. Quantitative evaluation shows this approach significantly outperforms conventional 2D full-range baselines, achieving a 6.2\% improvement in FID and superior MMD, Precision, and Recall across all HU intervals. The best performance is achieved by a multi-head VQVAE variant, demonstrating that it is possible to enhance visual fidelity and variability while also reducing model complexity and computational cost. This work establishes a new paradigm for structure-aware medical image synthesis, aligning generative modelling with clinical interpretation.
    
\end{abstract}


\section{Introduction}
\label{sec:background}

    \subsection{CT Scans and Hounsfield Units}

    
    Lung cancer is one of the deadliest cancers worldwide, accounting for the highest rate of cancer-related deaths (approximately 18.7\%) both in men and women, in 2022 \citep{lung-cancer-statistics}. Computerized Tomography (CT) chest scans have been adopted as the main tool for diagnosing lung cancer, and other lung diseases, as they provide the ability to visualise pulmonary anatomy and assess the presence of tumours, their exact location, size and stage with accurate diagnostic images, playing a key roll for patient diagnosis and treatment planning \citep{ct-scans-cancer-1, ct-scans-cancer-2}.
    
    CT scan machines employ X-ray detectors and beams that rotate around a patient's body to capture 2D or 3D cross-sectional images of the region of interest for the specific diagnosis. During image reconstruction, each tissue's X-ray absorption coefficient are quantified into Hounsfield units (HU), computed from a linear transformation of the baseline attenuation coefficient of the X-ray beam \citep{hounsfiled-unit}, ranging from $-1000$ to $3000$. The greater the tissue density, the higher the absorption coefficient is, and, therefore, the higher the HU value will be. Table \ref{tab:hu-tissue-ranges} frames the HU values of some tissues and materials found in CT scans.
    
    \begin{table}[htb!]
        \centering
        \caption{Hounsfield units for different tissues and materials \citep{hu-table-values}}
        {\scriptsize
        \begin{tabular}{ll}
            \hline
            \textbf{Tissue} & \textbf{Hounsfield Unit (HU)}\\ \hline
            Air & $-1000$\\ \hline
            Lung & $[-700, -600]$\\ \hline
            Fat & $[-120, -90]$\\ \hline
            Water & $0$\\ \hline
            Abscess/pus & $0$ or $20$, $40$ or $45$\\ \hline
            Blood & $[30, 45]$\\ \hline
            Muscle & $[35, 55]$\\ \hline
            Bone & $[700, 3000]$\\ \hline
        \end{tabular}
        }
        \label{tab:hu-tissue-ranges}
    \end{table}


    \subsection{Generative Approaches for Medical Imaging}
    
    The use of Artificial Intelligence (AI), in particular Deep Learning, has been extensively explored for applications in Computer Assisted Diagnosis (CAD) systems, such as nodule segmentation and classification in lung cancer screening. However, data availability limitations impose challenges for model training \citep{data-availability}, while the curation process of such data collections also raises new challenges due to time and costs, as well as data validation from medical teams \citep{data-augmentation-challenges}, to ensure any trained model has processes clinical trustworthy data. 
    
    To address this challenge, Image Generative AI models have surfaced as a promising solution \citep{data-augmentation-challenges}, capable of producing synthetic data with realistic and structural and anatomical characteristics derived from training data. In recent years, many Generative AI has been applied to the context of medical images, such as in image enhancement for noise reduction \citep{ct-scans-cancer-1}, super-resolution \citep{dlsuperresolution} and image-to-image modality translation \citep{image-translation}.
    
    As of late, many generative methods employ techniques such as Generative Adversarial Networks (GANs), Diffusion Models (DMs) and corresponding score-based implementations, and, latent generative models, for example Vector-Quantized Variational Autoencoders (VQVAEs).


        \subsubsection{Generative Adversarial Networks}
        
\begin{figure}[tb]
            \centering
            \includegraphics[width=0.5\linewidth]{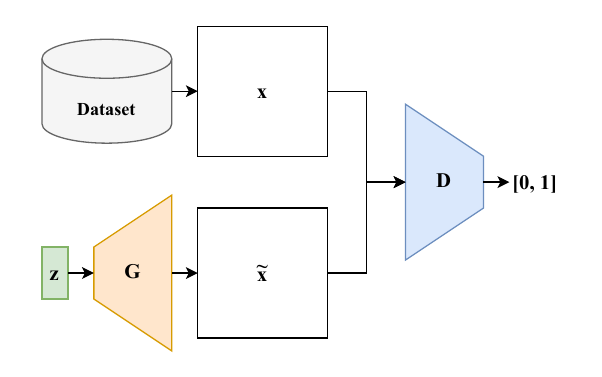}
            \caption{General GAN framework, where the Discriminator $D$ predicts the probability of authenticity for both real images from the dataset or synthetic ones, generated from the output of the Generator $G$ given a latent vector $z$.}
            \label{fig:gan-framework}
        \end{figure}

        GANs formulate generative modelling as a two-player minimax game between a generator network $G$ and a discriminator network $D$ \citep{gans}. The generator maps samples from a prior distribution $z \sim p_z$ to the data space, $G(z)$, while the discriminator distinguishes between real samples $x \sim p_{data}$ and generated ones $\tilde{x} \sim p_G$. The general GAN framework is summarized on the diagram in Figure \ref{fig:gan-framework}. The standard GAN optimization objective is formulated by Equation \ref{eq:gan}.
        
        \begin{equation}
            \min_G \max_D \mathbb{E}_{x \sim p_{data}}[\log D(x)] + \mathbb{E}_{z \sim p_z}[\log (1-D(G(z)))]
            \label{eq:gan}
        \end{equation}
        
        Ideally, the adversarial game formulation converges when the $p_G$ matches $p_{data}$. However, while effective, GANs often suffer from unstable training and mode collapse due to vanishing gradients when $D$ strongly outperforms $G$. To address these issues, Wasserstein GANs (WGANs) \citep{wgans} reformulate the objective by minimizing the Earth Mover distance (Wasserstein distance) between $p_{data}$ and $p_G$, and restricting the family of functions of $D$ to the set of 1-Lipschitz functions, $\mathcal{D}$. The Lipschitz constraint is crucial, because it stabilizes training by encouraging the discriminator's gradients to have unit norm, leading to more robust convergence.
        
        Instead of enforcing the constraint through weight clipping, \citet{wgans-gp} proposed the WGAN with Gradient Penalty (WGAN-GP), which adds a soft penalty on the gradient norm to the optimization objective, framed in Equation \ref{eq:wgan-gp}, where $\bar{x}$ are samples interpolated between real and generated data.
        
        \begin{equation}
            \min_G \max_D \mathbb{E}_{x \sim p_{data}}[D(x)] - \mathbb{E}_{z \sim p_z}[D(G(z)))] + \lambda_{GP} \cdot \mathbb{E}_{\bar{x} \sim p_{\bar{x}}} (||\nabla_{\bar{x}} D(\bar{x})||_2 - 1)^2
            \label{eq:wgan-gp}
        \end{equation}


        \subsubsection{Score-based Diffusion Models}
        
\begin{figure}[tb]
            \centering
            \includegraphics[width=0.85\linewidth]{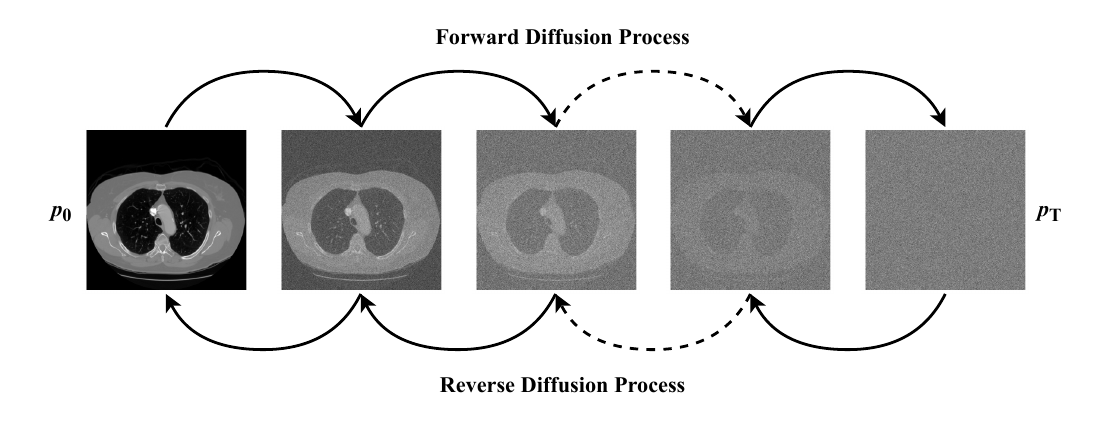}
            \caption{Effect of forward and reverse diffusion processes on a chest CT scan slice.}
            \label{fig:ddpm-noise-schedule}
        \end{figure}

        Denoising Diffusion Probabilistic Models (DDPMs) \citep{ddpm} operate by first defining a forward process that gradually perturbs data from the original distribution $p_0(\mathbf{x})$, with Gaussian noise over a fixed number of $T$ timesteps, until it approaches a pure Gaussian prior distribution $p_T(\mathbf{x})$. Afterwards, it is possible to learn the reverse process, which iteratively denoises the samples step by step, allowing for the reconstructions of realistic data from pure noise. The effects of the forward and reverse processes are shown in Figure \ref{fig:ddpm-noise-schedule}.

        This discrete formulation can be generalized to a continuous-time approach, where score-based generative modelling defines the reverse process as the solution of a Stochastic Differential Equation (SDE). Letting $\{\mathbf{x}(t)\}_{t \in  [0, T]}$ represent a continuous diffusion process, the corresponding stochastic forward and reverse processes are defined by Equation \ref{eq:forward-sde} and \ref{eq:reverse-sde}, respectively.
        
        \begin{equation}
            \mathrm{d}\mathbf{x} = \mathbf{f}(\mathbf{x}, t) \mathrm{d}t + g(t) \mathrm{d}\mathbf{w}
            \label{eq:forward-sde}
        \end{equation}
        \begin{equation}
            \mathrm{d}\mathbf{x} = [\mathbf{f}(\mathbf{x}, t) - g^{2}(t)  \nabla_{\mathbf{x}} \log p_t(\mathbf{x})] \mathrm{d}t + g(t) \mathrm{d}\mathbf{\bar{w}}
            \label{eq:reverse-sde}
        \end{equation}
        
        Next, setting the \textit{drift} coefficient $\mathbf{f}(\cdot, t)$ and the \textit{diffusion} coefficient $g(t)$ as in Equation \ref{eq:vp-sde} describes a \textit{variance preserving} SDE that equivalently models DDPM's discrete diffusion process \citep{score-based-dms}, where $\beta(t)$ describes a linear noise schedule.
        \begin{equation}
            \mathbf{f}(\mathbf{x}, t) = -\frac{1}{2}\beta(t)\mathbf{x} ,\quad g(t) = \sqrt{\beta(t)}, \quad \beta(t) = (\beta_{max} - \beta_{min}) \frac{t}{T} + \beta_{min}
            \label{eq:vp-sde}
        \end{equation}
        
        Finally, through the optimization of a time-informed neural network to estimate the score-function of each timestep marginal distribution, $\nabla_{\mathbf{x}} \log p_t(\mathbf{x})$, it is possible to sample data from $p_0(\mathbf{x})$ by executing the reverse diffusion process from Gaussian noise. The score-predicting model's architectures are usually time-informed U-Nets, composed by an downsampling, encoding component, an upsampling, decoding component, and skip connections for the use of shared information from earlier layers, alongside embedding-producing shallow multi-layer perceptrons to integrate timestep information for score prediction. The optimization of a timestep-informed score-based model $s$ is performed over the minimization of MSE to the score-function through denoising score matching \citep{denoising-score-matching}, as displayed in Equation \ref{eq:dm-optimization}, where $\lambda$ is an appropriate positive weighting function \cite{song2021maximum} and $t$ is uniformly sampled over $[0, T]$.
        
        \begin{equation}
            \min \mathbb{E}_{t} \bigg\{ \lambda(t) \mathbb{E}_{\mathbf{x}(0)} \mathbb{E}_{\mathbf{x}(t) | \mathbf{x}(0)} \big[ || s(\mathbf{x}(t), t) - \nabla_{\mathbf{x}(t)} \log p_{0t}(\mathbf{x}(t) | \mathbf{x}(0)) ||_{2}^{2} \big] \bigg\}
            \label{eq:dm-optimization}
        \end{equation}


        \subsubsection{Vector-Quantized Variational Autoencoders}
        
        \textit{Two-stage approaches} family of generative models attempt at learning encoded representations of the data distribution, and later optimize a generative model to learn the encoded representation distribution. Some works have shown state-of-the-art results using these approaches, such as Latent Diffusion Models \citep{ldms}, Variational Diffusion Models \citep{variational-dms}, as well as VQVAEs \citep{vqvaes} and VQGANs \citep{vqgans}. Learning meaningful and compact latent representations, capturing the essential semantic structures of the data, is needed to, consequently, operate more tractably, efficiently and stably in a reduced-dimensional latent space during the generative modelling stage. While most encoding methods provide continuous features for generative modelling, VQVAEs focus on the extraction of discrete representations \citep{vqvaes}.
        
        On the first stage of training, the VQVAE learns an encoder model $E$ and decoder model $D$, such that, together, they learn to represent images using codes from a learned discrete codebook $\mathcal{Z} = \{z_c\}_{c=1}^C$, where $z_k \in \mathbb{R}^{n_z}$. More specifically, an image $x \in \mathbb{R}^{H \times W \times C}$ is approximated by $\hat{x} = D(z_{\mathbf{q}})$, where the quantized latent representation $z_{\mathbf{q}}$ is obtained from the encoder output $\hat{z} = E(x) \in \mathbb{R}^{H \times W \times n_z}$ through element-wise nearest-neighbour quantization, as in Equation \ref{eq:quantization}.
        
        \begin{equation}
            z_{\mathbf{q}} = \mathbf{q}(\hat{z}) = \arg \min_{z_c \in \mathcal{Z}} ||\hat{z}_{ij} - z_c||
            \label{eq:quantization}
        \end{equation}
        
        Thus, the reconstruction is given by $\hat{x} = D(z_{\mathbf{q}}) = D(\mathbf{q}(E(x)))$.
        
        Since the quantization operation in Equation \ref{eq:quantization} is non-differentiable, a straight-through gradient estimator \citep{gradient-estimator} is used to copy gradients from the decoder back to the encoder, allowing the training of both the codebook and the model. The overall training loss function is shown in Equation \ref{eq:vqvae-loss}, where the first term denotes the reconstruction loss, $\mathbf{sg[\cdot]}$ is the stop-gradient operation, and the last term is the commitment loss \citep{vqvaes} that encourages the encoder outputs to remain close to their corresponding codebook entries, with $\lambda_c$ being a weighting factor.
        
        \begin{equation}
            \mathcal{L}(E, D, \mathcal{Z}, x) = ||x - \hat{x}||_2^2 + ||\mathbf{sg}[E(x)] - z_{\mathbf{q}}||_2^2 + \lambda_{\text{c}}||\mathbf{sg}[z_{\mathbf{q}}] - E(x)||_2^2
            \label{eq:vqvae-loss}
        \end{equation}
        
        On the second training stage, given the trained encoder $E$ and decoder $D$, feed-forwarded images can be represented as sequences of codebook indices corresponding to their quantized latent encoding, i.e., the quantized latent representation of an image $z_{\mathbf{q}} \in \mathbb{R}^{H \times W \times n_z}$ is mapped to a sequence $s \in \{0, 1, ... C-1\}^{H \times W}$, where each element $s_{ij}$ indexes the its nearest codebook vector $z_c$. By converting the indices in $s$ back to their respective codebook entries, the image can be reconstructed as $\hat{x} = D(z_{s_{ij}})$.
        
        To model the data distribution, a transformer model is trained auto-regressively on the index sequences, predicting each token $s_i$ conditioned on all previous ones $s_{<i}$, to compute the likelihood of the full representation as $p(s) = \prod_i p(s_i | s_{<i})$. Thus, the model maximizes the log-likelihood of the discrete latent representations, defining the transformer loss function as in Equation \ref{eq:transformer-loss}.
        
        \begin{equation}
            \mathcal{L}_{\text{Transformer}} = \mathbb{E}_{x \sim p(x)} [- \log p(s)]
            \label{eq:transformer-loss}
        \end{equation}

\begin{figure}[tb]
            \centering
            \includegraphics[width=1.0\linewidth]{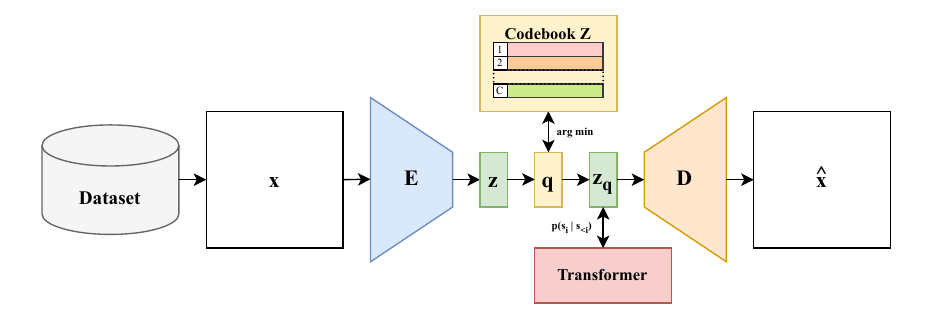}
            \caption{General VQVAE framework. A given image $x$ passes through the Encoder network $E$ to produce a continuous latent representation $z$. Next, given the codebook $\mathcal{Z}$ with $C$ vectors, the quantization operator $\mathbf{q}$ produces a discrete latent representation $z_{\mathbf{q}}$. Finally the decoder network $D$ predicts a reconstruction of the original image $\hat{x}$ by decoding $z_\mathbf{q}$. The transformer module auto-regressively generates sequences of discrete values that try to match those produced by $z_\mathbf{q}$, allowing the synthesis of new images.}
            \label{fig:vqvae-framework}
        \end{figure}

        A VQVAE generates new samples by generating sequences of codebook indexes, which are then converted to their vector representation and re-ordered back to grid structure, and, finally, fed through the decoder network to obtain the new, synthetic images. The overall VQVAE framework is displayed in the diagram of Figure \ref{fig:vqvae-framework}.

    \subsection{Motivation \& Contributions}

    In clinical practice, radiologists often adjust HU windows to better visualise specific tissues. Therefore, this windowing method can effectively perform a form of semantic segmentation for different anatomical structures in a totally analytical manner. Despite this being a routine practice, current generative approaches mostly ignore this form of decomposition, and attempt to deal with full-range CT scans as a single continuous distribution, seeing as this is the modality most useful in any real world application.

    In this work, we propose a two-step generative framework to decompose full CT generation into structure-specific subtasks. Rather than using traditionally full-domain scans, this allows for the joint modelling of various individual HU intervals, each one ideally corresponding to a distinct tissue. Their outputs are then fused through a simple reconstruction network, effectively imitating the reverse process of HU windowing with minimal information loss. Both training and validation are performed on the publicly available LIDC-IDRI dataset~\citep{lidc-1, lidc-2} and resorting to both quantitative key metrics and qualitative expert evaluation.

    
    This study therefore exploits a disconnect between existing generative literature and clinical interpretation of lung CT scans. The advantages brought about by this method are fourfold: by using multiple generators, each focused on a narrower intensity distribution, it is possible to reduce model complexity and optimise computational efficiency - especially relevant in resource-strained scenarios / environments. Moreover, this allows for better interpretability and more control over the results, seeing as clinicians are given access not only to the final full-range CT scans but also all the HU windows it is comprised of.
    
    In summary, the contributions of this research are as follows:

    \begin{itemize}

        \item Introduction of a novel HU-interval decomposition strategy for full lung CT scan generation, with the aim of aligning generative modelling and clinical practice and enabling a targeted, tissue-specific manner of generation.
        \item Proposal and comparison of three architectural variants for a HU window-aware generative design, with the multi-head VQVAE being the one to achieve a better performance.
        \item Enhancement of generative explainability through the use of a transparent stepwise process which enables clinicians to interpret and validate the synthetic output at a macro and component levels.
        \item Expansion upon the concept of HU windowing by providing a learned reconstruction method to reverse it with minimal loss of information.
        \item Demonstration of significant improvements in both perceptual quality and diversity over traditional AI-based full-range baselines, while also priming for a decrease in computational demands and training costs.
        \item Conduction of a Visual Turing Test (VTT) with practising clinicians with satisfactory results, thus providing real clinical validation by proving the results' anatomical realism and diagnostic relevance.
        \item Creation of an open framework that may possibly be extended to other medical imaging modalities where intensity-based decomposition is possible and meaningful.
        
    \end{itemize}

    The remainder of the manuscript is organised as follows: section~\ref{sec:related-work} introduces the related work and literature state-of-the-art on the matter; section~\ref{sec:methodology} presents our designed architecture and pipeline for decomposed full lung CT scan generation; section~\ref{sec:results-and-discussion} we discuss the extensive experimentation regiment conducted in order to evaluate method performance quantitatively and qualitatively; and section~\ref{sec:conclusion} finally concludes the paper with a brief overview on contributions and future potential.


\section{Related Work}
\label{sec:related-work}

    The field of medical image synthesis has evolved significantly over the years. Early analytical physics-based approaches relied mostly on mathematical phantoms~\citep{segars_4d_2010, shepp_fourier_1974} (e.g. the Shepp-Logan phantom) or finite-element simulations~\citep{werner_patient-specific_2009} which would produce images to be used in calibration and algorithm validation for the most part. These approaches saw variability being handled through the use of linear combinations for principal components, in what were called statistical shape and active appearance modelling~\citep{cootes_active_1995, cootes_active_2001}. Furthermore, atlas-based synthesis~\citep{iglesias_multi-atlas_2015} was a staple of multimodality, seeing as generation was done via registration to a template. This method, specifically allowed for improved interpretability and control over anatomical features, as well as facilitating downstream tasks, such as segmentation. However, analytical methods were inherently unable to fully capture and mimic the complexity of real samples, being limited by faulty physic assumptions and low-dimensional representations, even despite their reliance on experts defining priors.

    The rise of machine and deep learning shifted this paradigm from using simplified physics approximations to iterative data-driven modelling~\citep{dayarathna_deep_2024, patel_deep_2025}. This revolutionised not only the generative scene, but perhaps more importantly contributed greatly towards the optimisation and automation of downstream tasks such as pathology prediction~\citep{salama_generalized_2022}. Variational autoencoders (VAEs) were some of the first on the scene to be applied to not only representation learning of healthy lung CT scan~\citep{zimmerer_unsupervised_2019, cetin_attri-vae_2023, varma_medvae_2025, rais_exploring_2024} but also for the purpose of enhancing lung nodule variability~\citep{liu_attention-based_2023, li_3d_2024, patel_enhancing_2025}. It should also be denoted how, in the same sense, convolutional neural networks (CNNs) also demonstrated the potential to produce full high-resolution CT scans in a number of different works and tasks, including pure synthesis~\citep{nie_medical_2018}, imaging modality translation~\citep{han_mr-based_2017}, etc. Some major examples of such an application include the works of Chen et al.~\cite{chen_unsupervised_2020} who employed a VAE to create synthetic lung CT patches as a way of constructing a continuous texture latent space and more recently Singla et al.~\citep{singla_data_2025}, whose use of VQVAEs for this data further prove the still-existing capabilities of autoencoders in the generative scene. Together, CNNs and VAEs can be considered the first deep learning applications to CT data augmentation and, therefore, the ones to lay the foundation for lung anatomy representation learning, despite their tendency to produce blurry and/or detail-lacking samples.

    This shift in data augmentation was further pushed by GANs, whose ability to produce samples with strong perceptual quality and realistic textures far surpassed that of autoencoders and CNNs~\citep{chen_deep_2018, shin_medical_2018}. The adoption of this new adversarial golden standard quickly proved to cause a significant leap, marked by an explosion in the number of tasks for which generative models could be adapted to within the medical imaging field~\citep{skandarani_gans_2021, kazeminia_gans_2020}, and more specifically regarding lung CT data~\citep{gonzalez-abril_statistical_2022, astley_deep_2022}. Consequently, there are a variety of works which have employed GANs trained on lung CT scan data for tasks such as denoising~\citep{yang_low-dose_2018, wolterink_generative_2017} and more importantly data augmentation. For instance, both Chuquicusma et al.~\citep{chuquicusma_how_2018} and Jin et al.~\citep{jin_ct-realistic_2018} demonstrated how a GAN-based model can produce fake lung nodules realistic enough to deceive radiologists into thinking they are real; Han et al.~\citep{han_synthesizing_2019} and Mendes et al.~\citep{mendes_lung_2023} developed a conditional GAN framework for the creation of full 2D lung CT slices using semantic anatomical maps; Salehinejad et al.~\citep{salehinejad_synthesizing_2019} utilised a deep convolutional GAN to generate high-resolution samples with synthetic interstitial pathological visual markers. The COVID-19 pandemic in 2020 also significantly increased the need for synthetic data for the sake of expediting the screening and diagnosis processes, thus prompting numerous GAN-based augmentation efforts~\citep{li_covid-19_2021, menon_ccs-gan_2023, loey_within_2020, kaur_synthetic_2023}. This was due to the GAN framework's versatility, seeing as its main strength is how it can combine its adversarial training nature with any other generative model, including VAEs. This is not, however, without drawbacks, as GANs are very data-sensitive~\citep{ellis_evaluation_2022} and prone to a phenomenon known as mode collapse~\citep{thanh-tung_catastrophic_2020}.

    Mostly though, early GAN approaches also lacked precise control over structural coherence and especially the fine detail intricacies of lung CT scans. As such, the newest state-of-the-art benchmarks in output quality and variability have been set by DDPMs, which traded the single-pass generation process used by GANs for an iterative denoising process with far superior results both in the understanding of underlying distributions and in the training stability~\cite{shi_diffusion_2025, akbar_beware_2024}. A foundational and major advancement in 3D medical imaging generation was established by Khader et al.~\citep{khader_denoising_2023}, who proved the success in applying this architecture for volumetric data and bypassed significant memory and complexity challenges to obtain wondrous results. Within the field of lung CT scan production specifically, diffusion models have been widely adopted: Jiang et al.~\citep{jiang_lung-ddpm_2025} optimised an efficient diffusion model for thoracic CT scans by balancing perceptual quality and reduced computational expenses; Pan et al.~\cite{pan_2d_2023} focused on displaying the scalability of transformer-based diffusion models for 2D synthesis, having achieved high fidelity slice-wise generation; Jiang et al.~\citep{jiang_fast-ddpm_2025} utilised a Fast-DDPM for image-image translation, effectively proving that diffusion models can be a viable option for time-sensitive applications and real time inference; Chen et al.~\citep{chen_cbct-based_2024} who also addressed modality translation from cone-beam CT (CBCT) scans to regular CT images, eliminating the need for accurate dose calculation when performing CBCT radiotherapy; Daum et al.~\cite{daum_differentially_2024} extended diffusion into the clinically relevant field of privacy protection by developing a differentiable 3D latent diffusion framework for secure medical image generation. Beyond unconditional synthesis and modality translation, diffusion models have also shown strong performance in conditional / pathology-aware generation as seen in the works by Kaur et al.~\cite{kaur_synthetic_2023} on COVID-19 chest X-ray generation, Zhang et al.~\citep{zhang_gh-ddm_2023} on multi scale conditioning and Zhao et al.~\citep{zhao_4d_2025} on visual modelling of CT scan disease progression. One other notable work in this same line is that of Olivera et al.~\citep{oliveras_land_2025} which succeeds at producing high-fidelity 3D chest CT volumes with explicit control over both global anatomy and local nodule features, thus marking a significant step forward towards clinically-relevant conditionally-aware 3D imaging synthesis. Finally, although these methods eliminate the threats of mode collapse, unstable training and imprecise control over fine-grained anatomical details, they suffer mostly from slower inference, which is, as shown by some of the works cited, a downside current research is actively working towards resolving.


    However, more than being solely dependant on architecture and pipeline choice, advancing the field of medical imaging synthesis requires moving towards clinical alignment and reflecting the workflow and priorities of radiological practice. One such example is provided by Tang et al.~\citep{tang_disentangled_2021}, who use disentangled representation learning to decompose pathological manifestation in chest X-rays from underlying anatomy. Similarly, Cao et al.~\cite{cao_art-asyn_2025} developed an anatomy-aware texture generation framework to effectively handle and control the synthesis of illness-caused anomalies separately from the image's anatomical structure. And so, even though this pursuit for interpretability and control has manifested in a number of different works, none make an attempt at the task of generation while building upon HU-based preprocessing, which as shown by Hu et al.~\cite{hu_automatic_2001} is a foundational cornerstone of radiology workflow.
    
    Despite these advances, a common methodological limitation persists: most works handle lung CT data as if its intensity distribution is homogeneous, and can be modelled in its entirety without the need to address specific components separately. This, however, is contrary to radiology practice, where HU windowing is usually used to isolate and evidence specific intensity ranges / tissue types~\cite{hu_automatic_2001}. This oversight, creates a discrepancy between the synthesis process and radiologist's interpretation of the data, which may result in harder acceptance of the technology and its output. It is also true that using single network frameworks to learn complex distributions of data that can be decomposed could be considered an inefficient use of resources, especially in environments where these are spread thin. Only one other prior work has explicitly decomposed the generation of this data type according to this clinically grounded HU windowing practice, although, in their work, Krishna et al.~\citep{krishna_image_2024} depend on the existence of segmentation maps as a way to guide the generative process, which are not easily available and require the execution of unnecessary prior downstream tasks. To the best of our knowledge, this paper is the first to address these gaps by introducing a truly unbound generative HU-interval decomposition strategy which aligns generation with clinical workflow, while also improving upon the efficiency and output fidelity of lung CT synthesis.


\section{Methodology}
\label{sec:methodology}

    \subsection{Full-range CT Reconstruction from HU Intervals}
    \label{sec:ct-reconstruction}
    
    Let $X : \mathcal{V} \rightarrow H \subset \mathbb{R}$ denote a CT scan as a function over the voxel domain $\mathcal{V} \subset \mathbb{Z}^3$, where each voxel position corresponds to a HU value within the full HU range $H$ of the scan. Then, consider a set of non-overlapping HU intervals, whose union does not necessarily cover the entire HU range of $X$.
    \begin{equation}
        \mathcal{I} = \{ I_1, I_2, \dots, I_K \}, \quad I_k = [\text{hu}_{min}, \text{hu}_{max}], \quad I_i \cap I_j = \varnothing
    \end{equation}
    
    For each interval $I_k \in \mathcal{I}$, $X_k$ is defined as the result from a clipping transformation that retains the HU value of $X$ within $I_k$ while setting values lower than $\text{hu}_{min}$ to $\text{hu}_{min}$ and values greater than $\text{hu}_{max}$ to $\text{hu}_{max}$. Afterwards, both the original scan and the interval-clipped scans are min-max scaled to the unit range $[0, 1]$. Thus, the set $\{ X_1, X_2, \dots, X_k \}$ represents multiple complementary views of the same CT scan, each capturing a limited portion of its original HU distribution.
    \begin{equation}
        X_k(v) = \frac{\text{clip}(X(v), \text{hu}_{min}, \text{hu}_{max}) - \text{hu}_{min}}{\text{hu}_{max} - \text{hu}_{min}}, \quad \text{clip}(h, a, b) = \left\{ \begin{array}{lll}
                a, & \mbox{if} & h < a \\
                h, & \mbox{if} & h \in [a, b] \\
                b, & \mbox{if} & h > b. \\
                \end{array}\right.
    \end{equation}
    
    The first key objective of this study is to demonstrate that it is possible to reconstruct the full-range scan $X$ from the collection of interval-clipped scans $\{X_k\}_{k=1}^K$ using a learned reconstruction model $\mathcal{R}$.

    Firstly, it should be noted that it is not possible to reconstruct $X$ from $\{X_k\}$ using a fixed, linear or affine combination. In particular, there exists no coefficients $\{\alpha_k\}$ and bias $\beta$ that satisfy
    \begin{equation}
        X(v) = \sum_{k=1}^{K} \alpha_k X_k(v) + \beta, \quad \forall v \in \mathcal{V}.
    \end{equation}
    This impossibility arises because each $X_k$ encodes a mapping from HU values to normalized intensities that are locally linear within $I_k$ but globally inconsistent across the whole HU range. Furthermore, a voxel's contribution to the reconstruction depends on which HU interval it originally belonged to, which varies across the image. To formalize this, it is possible to define binary masks that identify the active interval for each voxel as
    \begin{equation}
        M_k(v) = \left\{ \begin{array}{ll}
                1, & \mbox{if} \quad X_k(v) \in I_k \\
                0, & \mbox{otherwise.}
                \end{array}\right.
    \end{equation}

    The scaled full-range scan can then be expressed as a piecewise mapping
    \begin{equation}
        X(v) = \sum_{k=1}^{K} M_k(\phi_k(v)) \cdot \sigma(\phi_k(v)),
    \end{equation}
    where each composite function $\sigma(\phi_k(\cdot))$ reprojects a HU intensity of $X_k$ to the respective intensity in $X$. One can unfold the composition and describe $\phi_k(\cdot)$ as a transformation from the HU of $X_k$ to its original value in $I_k$, and $\sigma(\cdot)$ as a linear scaling function within the full-range of $X$. However, if the union of HU intervals does not cover the whole CT scan HU range, then those HU values must still be deductible from $\{X_k\}$.

    In this context, a reconstruction model $\mathcal{R}$, defined by a deep neural network, can be trained to estimate reconstruction of $X$ from $\{X_k\}$ by implicitly learning to approximate the non-linear mask functions $M_k(\cdot)$, corresponding reprojection functions $\phi_k(\cdot)$, as well as to model the reconstruction of HU values uncovered by $\mathcal{I}$. This work explores the use of Multi-layer Perceptrons (MLPs) and CNNs of different feature sizes and depths to model the reconstruction of full-range CT scans given a collection of HU-clipped views of those same scans.

\begin{figure}[tb]
        \centering
    
        \begin{subfigure}{0.75\textwidth}
            \centering
            \includegraphics[width=\linewidth]{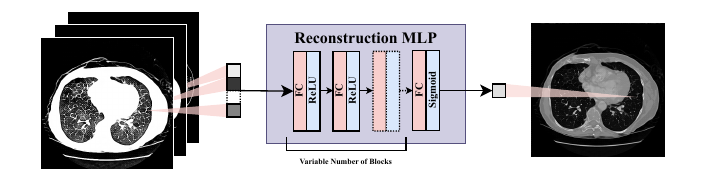}
            \subcaption{}
            \label{fig:mlp}
        \end{subfigure}
        
        \begin{subfigure}{0.75\textwidth}
            \centering
            \includegraphics[width=\linewidth]{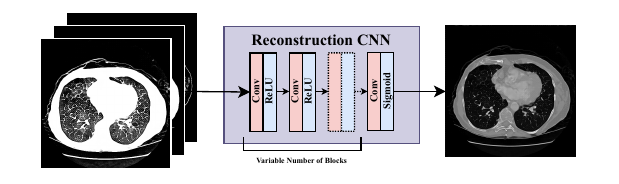}
            \subcaption{}
            \label{fig:cnn}
        \end{subfigure}
    
        \caption{Illustration of the proposed CT reconstruction networks. (a) An MLP-based model that receives, for each voxel, a vector of clipped-scaled HU values sampled from the corresponding voxel location across multiple input slices, and predicts the scaled full-range HU value using a sequence of fully connected–ReLU blocks followed by a sigmoid output layer. (b) A CNN-based model that takes a K-channel image in which each channel encodes a clipped-scaled HU slice, and reconstructs the corresponding scaled full-range CT image through a series of convolution–ReLU blocks and a final convolutional layer with sigmoid activation.}
        \label{fig:reconstruction-mlp-cnn}
    \end{figure}

    The MLP models receive a vector $\mathbf{x}_v = [X_1(v), X_2(v), \dots, X_K(v)]^T \in [0, 1]^K$, representing the $K$ scaled HU-window values at voxel $v$, and output a single scalar $\hat{X}(v) \in [0, 1]$. As depicted in Figure \ref{fig:reconstruction-mlp-cnn}\subref{fig:mlp}, the MLP architectures consist of an input layer of size $K$ followed by a variable number of full-connected (FC) layers, each also with feature size $K$. All hidden layers employ the ReLU activation function to introduce non-linearity, while the final output layer uses a Sigmoid activation to constrain the reconstructed voxel intensity to the $[0, 1]$ range. By processing all voxels of the given images, the full-range CT reconstruction estimation is obtained.

    On the other hand, in contrast to the voxel-wise MLP networks, the CNN models operate at the image level. Given $K$ input images $\{X_1, X_2, \dots, X_K\}$, each representing a scaled HU interval of a CT scan, the CNN receives a $K$-channel input tensor and predicts a single-channel output image $\hat{X}$, corresponding to the scaled, reconstructed full-range CT. The architecture is composed of a sequence of convolutional layers, each followed by a ReLU activation function. All convolutional layers preserve both spatial resolution and the number of channels, matching the dimensionality of the input, except for the final convolutional layer, which reduces the feature dimension to a single channel. The final output layer employs a Sigmoid activation to ensure that the reconstructed HU values remain within the scaled range $[0, 1]$. A diagram of the CNN architectures is also displayed in Figure \ref{fig:reconstruction-mlp-cnn}\subref{fig:cnn}.

    \subsection{Multiple HU Interval Generation}
    The second core goal of this study lies in the synthesis of multiple HU interval representations that enable the generation of high-quality, realistic, full-range CT samples. Instead of training a model to directly generate full-range CT scans, the proposed methods learn to generate several HU interval representations, each reflecting a specific sub-range of tissue or material contrasts, such that their combination through a reconstruction model produces a realistic, full-range CT sample.
    
    Hence, the presented work aims at learning a generative model that produces a sequence sample $\{\tilde{X}_1, \tilde{X}_2, \dots, \tilde{X}_K\}$ such that their reconstruction samples $\tilde{X} = \mathcal{R}(\{\tilde{X}_k\})$ create a full-range CT scan sample distribution $p_{\tilde{X}}$ similar to the original one $p_X$. This objective was explored using three generative paradigms, WGAN-GP, score-based DMs, and VQVAEs, each adapted to handle multi-interval generation through different architectural variants.

        \subsubsection{Multi-channel Approaches}
        The multi-channel modelling approach represents the most direct extension of traditional generative architectures to the multi-interval CT reconstruction setting. Its core idea lies in adapting single-channel generative frameworks to produce $K$-channel outputs, where each channel corresponds to a specific HU interval sample. This approach requires minimal modification to the original model architectures, therefore preserving inductive biases and optimization properties of the adapted framework. Since this modification can be applied seamlessly across diverse generative families, this multi-channel formulation can be regarded as a model-agnostic strategy for multiple HU interval generation.

\begin{figure}[tb]
            \centering
            \includegraphics[width=0.85\linewidth]{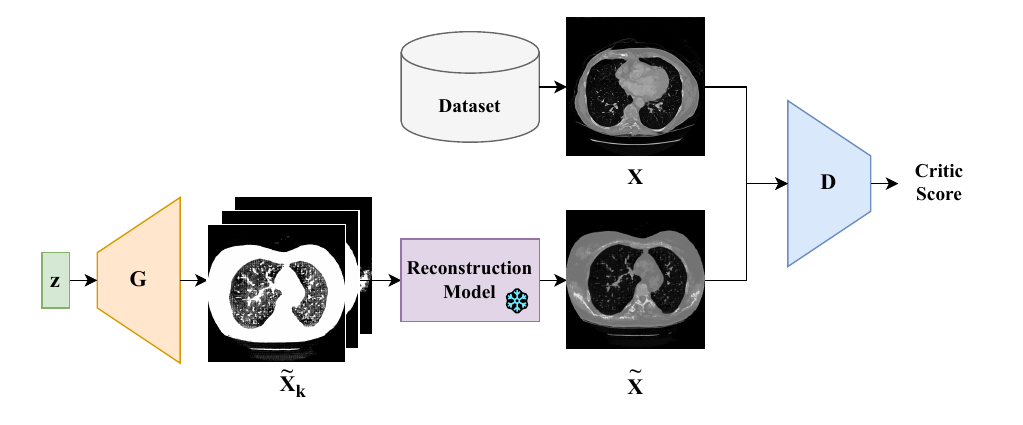}
            \caption{Multi-channel WGAN-GP framework.}
            \label{fig:multi-channel-gan}
        \end{figure}

        In the WGAN-GP configuration, as pictured in Figure \ref{fig:multi-channel-gan}, the generator network $G$ is adapted to produce a K-channel output, which is then passed through the reconstruction model $\mathcal{R}$ to obtain the final single-channel sample. The reconstructed images are inputted to the critic network $D$, which compares it against real full-range CT images.

\begin{figure}[tb]
            \centering
            \includegraphics[width=0.95\linewidth]{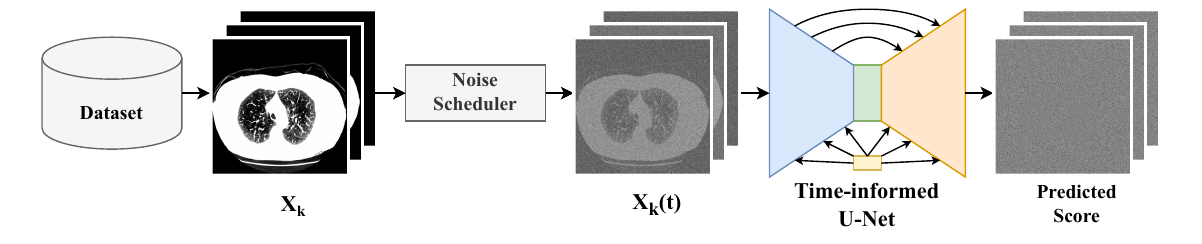}
            \caption{Multi-channel score-based diffusion model framework.}
            \label{fig:multi-channel-dm}
        \end{figure}

        For score-based DMs, the same principle applies, but the underlying architecture differs from the GAN framework. Here, a standard U-Net is employed to receive K-channel images and predict K-channel score values, displayed in Figure \ref{fig:multi-channel-dm}, thus directly modelling the denoising process in the multi-interval domain. Since the diffusion models in this work do not operate on direct image-prediction, the reconstruction model $\mathcal{R}$ is not directly used during training. After sampling, the $K$-channel HU intervals samples are processed by the reconstruction model $\mathcal{R}$ to obtain the full-range CT sample.

\begin{figure}[tb]
            \centering
            \includegraphics[width=1.0\linewidth]{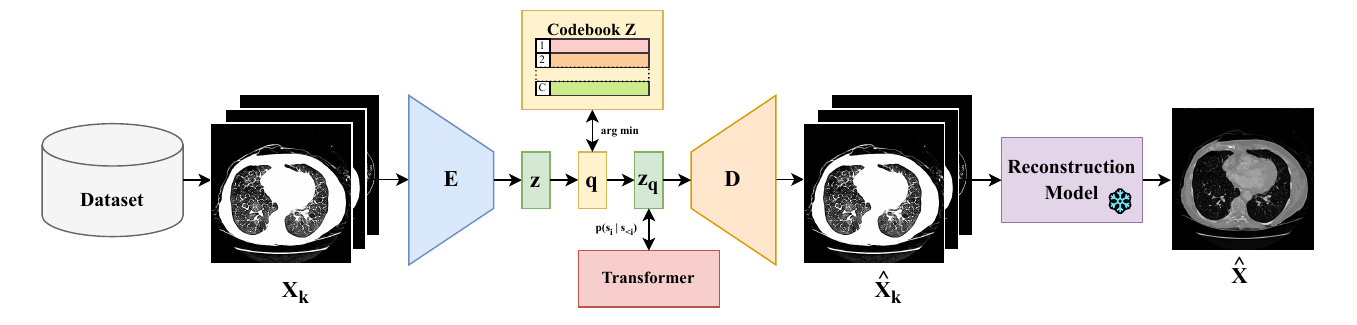}
            \caption{Multi-channel VQVAE framework.}
            \label{fig:multi-channel-vqvae}
        \end{figure}

        Finally, in the VQVAE framework, the encoder-decoder architecture is modified to operate over $K$-channel inputs and outputs, similarly to the previous generative approaches. The encoder $E$ compresses the multi-interval input into a shared latent codebook representation, while the decoder $D$ reconstructs the original $K$-channel image from the quantized latent embeddings. The multi-channel framework for VQVAE is framed in Figure \ref{fig:multi-channel-vqvae}. This adaptation pushes the encoder to model shared representations of the various HU intervals on each codebook entry. After training, the transformer model produces the quantized latent sequence, which decoded by $D$ produces the HU samples. Lastly, these are given to the reconstruction model $\mathcal{R}$ to produce the full-range CT synthetic image.

        \subsubsection{Multi-decoder and Multi-head Approaches}
        While the multi-channel strategy provides a simple and model-agnostic way of extending single-channel generative models to the multi-interval setting, it does not explicitly separate the modelling of individual HU intervals. In particular, all interval representations share the same encoding and decoding pathways, which may limit the model's ability to learn distinct statistical characteristics and textural details of specific HU ranges. To address this limitation, two alternative methods, multi-decoder and multi-head networks, were designed to provide finer modelling over interval-specific features. It is relevant to highlight that these approaches were implemented exclusively for the VQVAE generative framework, given that its two-stage structure divides the latent representation modelling and generative phases, offering greater flexibility for architectural modification in the encoding-decoding process.

\begin{figure}[tb]
            \centering
            \includegraphics[width=1.0\linewidth]{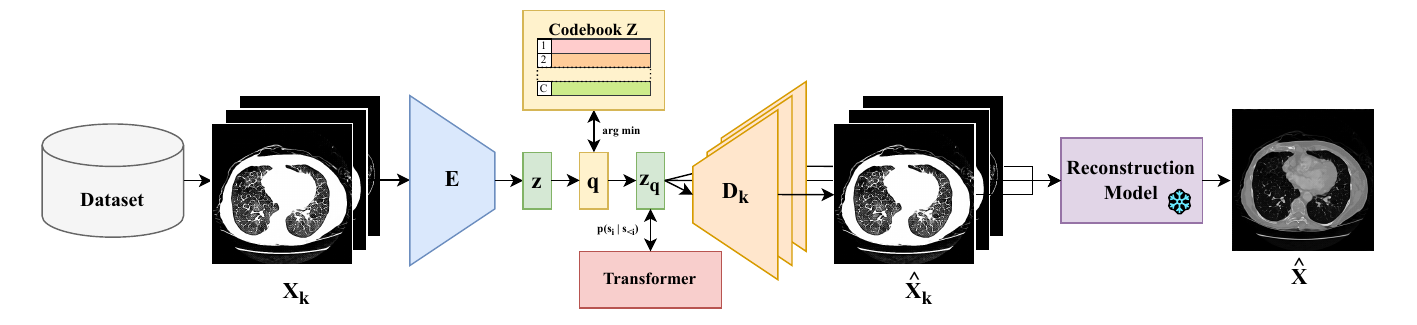}
            \caption{Multi-decoder VQVAE framework.}
            \label{fig:multi-decoder-vqvae}
        \end{figure}

        On one hand, the multi-decoder configuration is characterized by using a single encoder to process a concatenated, $K$-channel input representing the multiple HU clipped views, producing a unified latent representation that captures common structural information across HU intervals, as well as adjusting codebook entries to represent multiple textures for different HU content in the same quantized vector. Then, this representation is decoded independently by $K$ distinct decoder branches, each responsible for reconstructing one specific interval image. The outputs of these decoders are concatenated channel-wise and subsequently passed through the reconstruction model $\mathcal{R}$ to obtain the final full-range CT sample. The described architecture is displayed in Figure \ref{fig:multi-decoder-vqvae}. This approach encourages learning a shared latent representation, while allowing each decoder to specialize in each domain.

\begin{figure}[tb]
            \centering
            \includegraphics[width=1.0\linewidth]{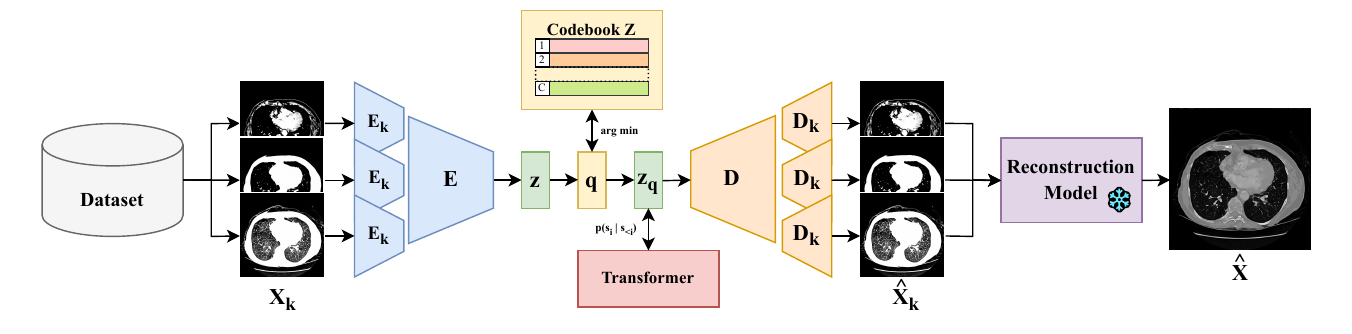}
            \caption{Multi-head VQVAE framework.}
            \label{fig:multi-head-vqvae}
        \end{figure}

        On the other hand, the multi-head method introduces interval-specific modelling at both the encoding and decoding stages. It employs $K$ separate encoder heads, each dedicated to processing one HU interval input, followed by a shared encoder-decoder backbone that performs mid-fusion of each intervals extracted features from their respective encoder head. Similarly to the other approaches, this one also pushes the codebook to produce entries representative of the various HU views values. The shared decoder path of the backbone is then followed by $K$ separate decoder heads that reconstruct the corresponding interval outputs. The resulting $K$ single-channel outputs are again concatenated and fed through the reconstruction model $\mathcal{R}$ to yield the full-range CT sample. An illustration of the multi-head architecture is framed in Figure \ref{fig:multi-head-vqvae}.

        Collectively, these VQVAE variant methods explore the trade-off between shared and specialized representations for multiple HU interval modelling in the same network.

        \subsubsection{Loss Function Formulations}
        For both the WGAN-GP and score-based DM configurations, the loss functions remain consistent with their respective single-channel counterparts. However, the VQVAE loss used in this work extends the original definitions. Please note that the pre-trained reconstruction model $\mathcal{R}$ is not updated or fine-tuned during any of the generative models training process. For the following formulations, let $B$ stand for the batch size from which the given loss is being computed.
        
        \paragraph{WGAN-GP Loss.} For WGAN-GP, the adversarial objective preserves its original formulation, where the Generator $G$ and Critic $D$ losses are described as in Equation \ref{eq:loss-wgan-generator} and \ref{eq:loss-wgan-critic}, respectively.
        \begin{equation}
            \mathcal{L}_G = \frac{1}{B} \sum_{b=1}^{B} \Big[ - D(\mathcal{R}(G(z^{(b)}) \Big], \quad z \sim \mathcal{N}(0, I)
            \label{eq:loss-wgan-generator}
        \end{equation}
        \begin{equation}
            \mathcal{L}_D = \frac{1}{B} \sum_{b=1}^{B} \Big[ D(\mathcal{R}(G(z^{(b)}))) - D(X^{(b)}) + \lambda_{GP} \cdot (||\nabla_{\bar{X}^{(b)}} D(\bar{X}^{(b)})||_2 - 1) \Big], \quad z \sim \mathcal{N}(0, I)
            \label{eq:loss-wgan-critic}
        \end{equation}
        
        \paragraph{Score-based DMs Loss.} The denoising score-matching objective for diffusion models is computed over all K channels simultaneously, following the original score-predicting scheme. Note that when the perturbation kernel is Gaussian with variance $\sigma(t)^2$, the score $\nabla_{\mathbf{x}(t)} \log p_{0t}(\mathbf{x}(t) | \mathbf{x}(0))$ has a defined closed form:
        \begin{equation}
            X(t) = X(0) + \sigma(t) \cdot \varepsilon, \quad \varepsilon \sim \mathcal{N}(0, I)
            \quad \Rightarrow \quad
            \nabla_{X(t)} \log p_{0t}(X(t) | X(0)) = \frac{X(t) - X(0)}{\sigma(t)^2} = - \frac{\varepsilon}{\sigma(t)}.
        \end{equation}
        Hence, the loss is defined as in Equation \ref{eq:loss-dm}.
        \begin{equation}
            \mathcal{L}_{DM} = \frac{1}{B} \sum_{b=1}^{B} \Big[ ||s(X(t^{(b)}), t^{(b)}) + \varepsilon^{(b)} / \sigma(t^{(b)})||_2^2 \Big] , \quad \varepsilon \sim \mathcal{N}(0, I)
            \label{eq:loss-dm}
        \end{equation}

        \paragraph{VQVAE Loss.} For all VQVAE methods considered in this work, the same loss function is used to train the autoencoder of the first training stage of the generative framework. The complete loss function is composed of three main loss terms: pre-reconstruction loss, post-reconstruction loss, and quantization loss.

        The first term is designed to optimize the synthesis of each individual HU interval before reconstruction to a full-range CT sample. For each interval $k$, the VQVAE decoder output $\tilde{X}_k$ is compared with its corresponding ground-truth $X_k$ using a weighted combination of three complementary losses. The mean squared error ($\mathcal{L}_{MSE}$) computes pixel-wise reconstruction. Then, the complementary of the structural similarity index ($\mathcal{L}_{SSIM}$) accounts for structural consistency. The last term employs a perceptual loss ($\mathcal{L}_{RIN}$) over feature vectors of $\tilde{X}_k$ and $X_k$, extracted from a pre-trained ResNet-50 on the RadImageNet (RIN) dataset \citep{rad-image-net}.

        These losses are aggregated per interval and subsequently combined across all HU channels using a set of dynamic HU weights, in order to balance the influence of the losses across channels. More specifically, intervals containing a higher proportion of non-zero values receive larger weights, while sparse channels retain a minimum weight. The HU weights $\{w_k\}$ of the HU interval images $\{X_k\}$ from a CT scan image are computed as in Equation \ref{eq:hu-weight}, which also guarantees that the sum of all weights amounts to one.
        \begin{equation}
            \{w_k\} = \text{softmax}\Big( \Big\{ \max \Big(\frac{\#[X_k > 0]}{|X_k|}, w_{\text{min}} \Big) , \quad k \in \{1, 2, \dots, K\} \Big\} \Big)
            \label{eq:hu-weight}
        \end{equation}

        Then, the pre-reconstruction loss $\mathcal{L}_{\text{Pre-Rec}}$ is defined as in Equation \ref{eq:loss-pre-rec}, where $\lambda_{\text{Pre-}MSE}$, $\lambda_{SSIM}$, and $\lambda_{\text{Pre-}RIN}$ are positive loss weights for each respective atomic loss term in the pre-reconstruction loss definition, $\mathcal{L}_{MSE}$, $\mathcal{L}_{SSIM}$, and $\mathcal{L}_{RIN}$.
        \begin{equation}
            \begin{aligned}
            \mathcal{L}_{\text{Pre-Rec}} = \frac{1}{B} \sum_{b=1}^{B} \sum_{k=1}^{K} w_k^{(b)} \cdot \Big[ & \lambda_{\text{Pre-}MSE} \cdot \mathcal{L}_{MSE}(X_k^{(b)}, \hat{X}_k^{(b)}) + \\ & \lambda_{SSIM} \cdot \mathcal{L}_{SSIM}(X_k^{(b)}, \hat{X}_k^{(b)}) + \lambda_{\text{Pre-}RIN} \cdot \mathcal{L}_{RIN}(X_k^{(b)}, \hat{X}_k^{(b)}) \Big]
            \end{aligned}
            \label{eq:loss-pre-rec}
        \end{equation}

        Afterwards, the second term ensures that the reconstruction from the generated HU interval samples outputted from the reconstruction model $\mathcal{R}$ yields realistic and coherent full-range CT sample. The post-reconstruction loss ($\mathcal{L}_{\text{Post-Rec}}$), described in Equation \ref{eq:loss-post-rec} is composed by the sum of $\mathcal{L}_{MSE}$ and $\mathcal{L}_{RIN}$ on a full-range CT image $X$ and its full-range reconstructed version from the encoding-decoding process $\hat{X} = \mathcal{R}(\{\hat{X}_k\})$. The inclusion of this post-reconstruction term encourages independently learned HU interval representations contribute synergistically to an high-fidelity CT reconstruction. The terms $\lambda_{\text{Post-}MSE}$, and $\lambda_{\text{Post-}RIN}$ are weight coefficients for the contribution of $\mathcal{L}_{MSE}$ and $\mathcal{L}_{RIN}$ in the final total loss.
        \begin{equation}
            \mathcal{L}_{\text{Post-Rec}} = \frac{1}{B} \sum_{b=1}^{B}  \Big[ \lambda_{\text{Post-}MSE} \cdot \mathcal{L}_{MSE}(X^{(b)}, \hat{X}^{(b)}) + \lambda_{\text{Post-}RIN} \cdot \mathcal{L}_{RIN}(X^{(b)}, \hat{X}^{(b)}) \Big]
            \label{eq:loss-post-rec}
        \end{equation}

        Lastly, the quantization loss ($\mathcal{L}_{VQ}$) composes the last term of the VQVAE loss formulation, corresponding to the standard commitment and codebook update objectives in the generative framework, as seen in Equation \ref{eq:loss-quant}. This term stabilizes the discrete latent space by minimizing the discrepancy between continuous encoder embeddings and their quantized versions. The term $\lambda_{VQ}$ regularizes the weight of the quantization loss for the global VQVAE loss function.
        \begin{equation}
            \mathcal{L}_{VQ} = \lambda_{VQ} \cdot \frac{1}{B} \sum_{b=1}^{B}  \Big[ ||\mathbf{sg}[E(X^{(b)})] - z_{\mathbf{q}}^{(b)}||_2^2 + ||\mathbf{sg}[z_{\mathbf{q}}^{(b)}] - E(X^{(b)})||_2^2 \Big]
            \label{eq:loss-quant}
        \end{equation}

        Collecting the terms above, the total loss function for the VQVAE networks used in this study is defined as:
        \begin{equation}
            \mathcal{L}_{VQVAE} = \mathcal{L}_{\text{Pre-Rec}} + \mathcal{L}_{\text{Post-Rec}} + \mathcal{L}_{VQ}.
            \label{eq:final-loss}
        \end{equation}

    \subsection{Dataset and Preprocessing}
    The dataset employed in this study was derived from the publicly accessible Lung Image Database Consortium (LIDC-IDRI) collection \citep{lidc-1, lidc-2}. This database is composed of 1,018 thoracic CT scans acquired for lung cancer screening, each represented as a 3D volume of size $S \times 512 \times 512$, where $S$ denotes the number of axial slices per scan. The dataset utilized adheres to the ethical approvals, ensured on the corresponding cited description paper. For this work, 80\% of the volumes were allocated to the training set, while the remaining 20\% where reserved for testing, ensuring that no slices from the same patient appear across both sets. Moreover, only slices corresponding to indices 30 through 90 of each CT volume were retained for analysis.

\begin{figure}[tb]
        \centering
        \includegraphics[width=1.0\linewidth]{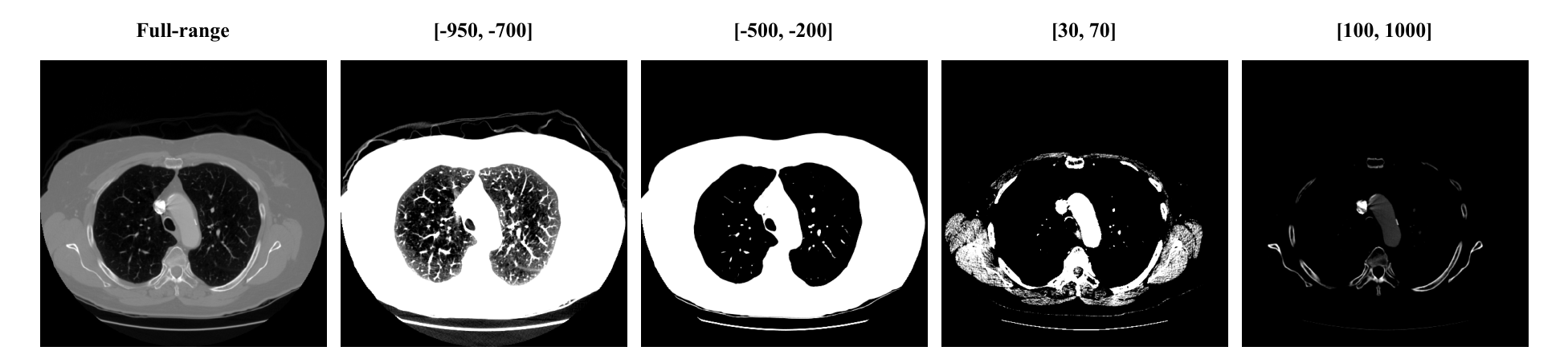}
        \caption{Example of a full-range CT sample and its respective HU-clipped views for the HU intervals $[-950, -700]$, $[-500, -200]$, $[30, 70]$ and $[100, 1000]$.}
        \label{fig:data-example}
    \end{figure}

    Each CT slice undergoes a preprocessing step consistent with the data preparation described in the beginning of Section \ref{sec:ct-reconstruction}. More specifically, a new preliminary step is performed, where the original CT slice is first clipped to the range $[-1000, 1000]$, and, afterwards, min-max scaled to $[0, 1]$, setting the full-range baseline representation. Next, for each HU interval in $\mathcal{I} = \{ [-950, -700], [-500, -200], [30, 70], [100, 1000] \}$, the HU interval representations were created using the same clipping and scaling procedure. Figure \ref{fig:data-example} displays an example of a full-range CT scan slice and its corresponding HU-windowed representations. The set of HU intervals previously defined was used throughout the entirety of the experiments of this work. The HU ranges were chosen considering the values in Table \ref{tab:hu-tissue-ranges}, and such that there is high independence between each HU views' textural details. The final dataset contains around 50 thousand data points consisted of a full-range CT scan slice and respective HU-clipped interval views, amounting to a total of nearly 250 thousand images in this work.

    \subsection{Experimental Setup}
    \label{sec:experimental-setup}
    Firstly, in the first set of experiments, it was evaluated the feasibility of reconstructing full-range CT scans from their respective HU-restricted representations using Deep Learning models. The network architectures employed for this objective are summarized in Table \ref{tab:ct-reconstruction-experiments}.
    
    \begin{table}[tb]
    \centering
    \caption{Configuration of the experimented reconstruction models}
    {\scriptsize
    \begin{tabular}{llll}
        \hline
        \textbf{Reconstruction Model} & \textbf{Model Type} & \textbf{Parameters} & \textbf{Description}                         \\ \hline
        MLP$_{0}$  & MLP & 5   & Zero hidden layers                   \\ \hline
        MLP$_{4}$  & MLP & 25  & One hidden layer with feature size 4 \\ \hline
        MLP$_{4 \times 4}$            & MLP                 & 45                  & Two hidden layers, all with feature size 4   \\ \hline
        MLP$_{4 \times 4 \times 4}$   & MLP                 & 65                  & Three hidden layers, all with feature size 4 \\ \hline
        CNN$_{3}$  & CNN & 37  & One convolution with kernel size 3   \\ \hline
        CNN$_{7}$  & CNN & 197 & One convolution with kernel size 7   \\ \hline
        CNN$_{11}$ & CNN & 485 & One convolution with kernel size 11  \\ \hline
        CNN$_{3 \times 3}$            & CNN                 & 185                 & Two convolutions with kernel size 3          \\ \hline
        CNN$_{3 \times 3 \times 3}$   & CNN                 & 333                 & Three convolutions with kernel size 3        \\ \hline
    \end{tabular}
    }
    \label{tab:ct-reconstruction-experiments}
    \end{table}
    
    All reconstruction models were trained and tested under the same conditions to ensure fair results comparison. Each model was optimized using a mean squared error (MSE) loss and trained with the Adam optimizer ($\beta_1 = 0.9$, $\beta_2 = 0.999$) with a learning rate of $5 \times 10^{-5}$, over 50 epochs with a batch of size 16. These hyperparameter values were chosen from empirical experimentation. The models performance was assessed using MSE, MS-SSIM, FID, MMD, Precision and Recall metrics, computed across 10 independent test runs, each involving the reconstruction of 256 full-range CT scans slices from their corresponding HU-clipped views.
     
    Next, for the second set of experiments, it was investigated whether the proposed methods for generating full-range CT scans via the reconstruction of synthetically generated HU-interval representations yield superior results compared to models trained to produce full-range images directly. Table \ref{tab:hu-generation-experiments} details the generative model types (WGAN-GP, Score-based DMs or VQVAEs), architectural variants and reconstruction models employed in each experimental configuration. The \textit{Baseline} approaches correspond to single-channel generative models trained directly on full-range CT images, serving as reference for evaluating the performance achieved by the proposed multi-interval generation methods.

    \begin{table}[tb]
    \centering
    \caption{Configuration of the experimented CT and HU generative models}
    {\scriptsize
    \begin{tabular}{llll}
        \hline
        \textbf{Model Type} & \textbf{Approach} & \textbf{Parameters (Millions)} & \textbf{Reconstruction Model} \\ \hline
        \multirow{3}{*}{WGAN-GP}        & Baseline                       & 6.4                   & -                           \\ \cline{2-4} 
                                        & \multirow{2}{*}{Multi-channel} & \multirow{2}{*}{6.8}  & MLP$_0$                     \\ \cline{4-4} 
                                        &                                &                       & CNN$_{3 \times 3 \times 3}$ \\ \hline
        \multirow{3}{*}{Score-based DM} & Baseline                       & 72.3                  & -                           \\ \cline{2-4} 
                                        & \multirow{2}{*}{Multi-channel} & \multirow{2}{*}{72.3} & MLP$_0$                     \\ \cline{4-4} 
                                        &                                &                       & CNN$_{3 \times 3 \times 3}$ \\ \hline
        \multirow{7}{*}{VQVAE}          & Baseline                       & 48.3                  & -                           \\ \cline{2-4} 
                                        & \multirow{2}{*}{Multi-channel} & \multirow{2}{*}{48.3} & MLP$_0$                     \\ \cline{4-4} 
                                        &                                &                       & CNN$_{3 \times 3 \times 3}$ \\ \cline{2-4} 
                                        & \multirow{2}{*}{Multi-decoder} & \multirow{2}{*}{68.9} & MLP$_0$                     \\ \cline{4-4} 
                                        &                                &                       & CNN$_{3 \times 3 \times 3}$ \\ \cline{2-4} 
                                        & \multirow{2}{*}{Multi-head}    & \multirow{2}{*}{49.3} & MLP$_0$                     \\ \cline{4-4} 
                                        &                                &                       & CNN$_{3 \times 3 \times 3}$ \\ \hline 
    \end{tabular}
    }
    \label{tab:hu-generation-experiments}
    \end{table}
        
    In the case of experiments using WGAN-GP models, the latent representations are 16-dimensional vectors sampled from Gaussian distribution, $\lambda_{GP}$ is set to 10, and the both the generator and critic networks use the Adam optimizer with beta parameters $\beta_1=0$ and $\beta_2 = 0.9$ with learning rate of $10^{-4}$. The training stage was executed for 20 epochs with batch size 16.

    Afterwards, for the experiments with Score-based DMs, the linear noise scheduler used $\beta_{min} = 0.1$ and $\beta_{max} = 20$ and 1000 timesteps for sampling, and the networks were optimized with Adam ($\beta_1=0.9$ and $\beta_2 = 0.999$) and learning rate of $5 \times 10^{-5}$ for 20 epochs with batch size 16. 

    After, for all experiments that employed VQVAEs, the corresponding codebooks were constituted by 512 vectors of dimensionality 16, and the commitment loss weight $\lambda_c$ was set to $0.5$. Additionally, the multi-decoder approach utilizes, on each decoding branch, the exact same decoder architecture of the multi-channel network, except for the number of output channels. Identically, the multi-head model uses the exact same encoder-decoder backbone architecture of the multi-channel approach, and only one encoder/decoder layer per input/output head, with the same characteristics of the corresponding blocks in the multi-channel network.
    
    Finally, the weights in the VQVAE loss formulation to optimize the autoencoder network were set as $\lambda_{\text{Pre-}MSE} = 1.0$, $\lambda_{SSIM} = 0.1$, $\lambda_{\text{Pre-}RIN} = 1.0$, $\lambda_{\text{Post-}MSE} = 0.1$, $\lambda_{\text{Post-}RIN} = 0.25$, and $\lambda_{VQ} = 1.0$. Moreover, the HU minimum weight in Equation \ref{eq:hu-weight} was set as $w_{\text{min}} = 0.15$. The first stage of training was conducted for 20 epochs with batch size 16, using the AdamW optimizer with beta parameters $\beta_1 = 0.9$ and $\beta_2 = 0.95$ and learning rate of $10^{-4}$. Secondly, the transformer network was optimized with same AdamW optimizer and beta values, although with learning rate of $2 \times 10^{-4}$, over 50 epochs with simulated batch size of 64 with accumulated gradients of micro batch size 16.

    It is highlighted that all hyperparameter values were obtained through empirical experimentation, where various configurations were tested across multiple runs to optimize model performance.

    Regarding performance evaluation of the all generative models, 5 distinct runs of testing were conducted to compute the average value of FID, MMD, Precision and Recall between 256 generated samples from each model and 256 sampled real CT images, for both full-range views and HU-interval ones. The MS-SSIM metric was also used in this evaluation procedure to extract a variety metric within the synthetic collection of images in each test run, by computing the average MS-SSIM value between each possible pair of images from the sample set.

    Moreover, it was employed an indirect task-based evaluation strategy to assess the fidelity and utility of the generated images from the multi-head VQVAE with CNN$_{3 \times 3 \times 3}$ pipeline. Specifically, using the lung-area segmentation model from \citep{segmentation-model}, the synthetic full-range CT samples are provided as input to the network and, given that the anatomical structure of the lung regions are well defined, the segmentation output provides an objective proxy for assessing whether the generated images preserve meaningful anatomical features. It is worth noting that this is a qualitative assessment, as there are no ground-truth data from the unconditionally generated images to produce quantitative values.

    \subsection{Performance Evaluation Metrics}
    To comprehensively assess the quality of reconstructed and generated CT slices, a combination of pixel-wise, distributional, and structural metrics is employed.

    Reconstruction voxel error is quantified using the MSE between generated and ground-truth CT slices for the reconstruction models experiments, providing a direct measure of voxel-level intensity deviation.

    Next, Fr\'{e}chet Inception Distance (FID) is reported to assess the alignment between real and generated data samples. In the case of this work, the features were extracted from the final average pooling layer of the ImageNet pre-trained \textit{InceptionV3} network, yielding 2048-dimensional vectors. Lower FID values indicate that the generated distribution more closely approximates the real data distribution in both mean and covariance, therefore reflecting higher perceptual fidelity and realism. Unlike FID, Maximum Mean Discrepancy (MMD) is a non-parametric measure of distributional divergence which does not assume Gaussian distribution of the feature space, making it complementary for assessing higher-order discrepancies. The MMD value is computed using the same \textit{InceptionV3} embeddings from FID, where smaller MMD values indicate greater similarity between the two distributions.

    Precision and Recall were adapted to assess the performance of generative image models by the proposed methodology of \citep{precision-recall}, which defines the fidelity and diversity of generated samples by comparing the manifold coverage of generated and real data distributions in a high-dimensional features space. For this study, the aforementioned embedding space used to obtain FID values was used as feature space, and the Precision and Recall values were obtained by implementing the algorithm provided by the cited work, where larger values indicate better performance. Precision measures the proportion of generated samples that lie within the support of the real data manifold, indicating the realism of the generated images, while Recall quantifies the fraction of real samples that are covered by the synthetic manifold, reflecting the generative diversity.

    Lastly, Multi-Scale Structural Similarity Index Measure (MS-SSIM) measures the structural similarity between two images across multiple spatial resolutions, combining luminance, contrast and structural comparisons at progressively coarser scales, yielding values in $[0, 1]$, with higher values indicating greater structural similarity. Furthermore, low MS-SSIM quantifies structural dissimilarity across multiple spatial scales and if computed pairwise among samples serves as a quantitative proxy for structural variability within a sample distribution.


\section{Results and Discussion}
\label{sec:results-and-discussion}

    \subsection{Full-range CT Reconstruction}
\begin{figure}[tb]
        \centering
        \includegraphics[width=1.0\linewidth]{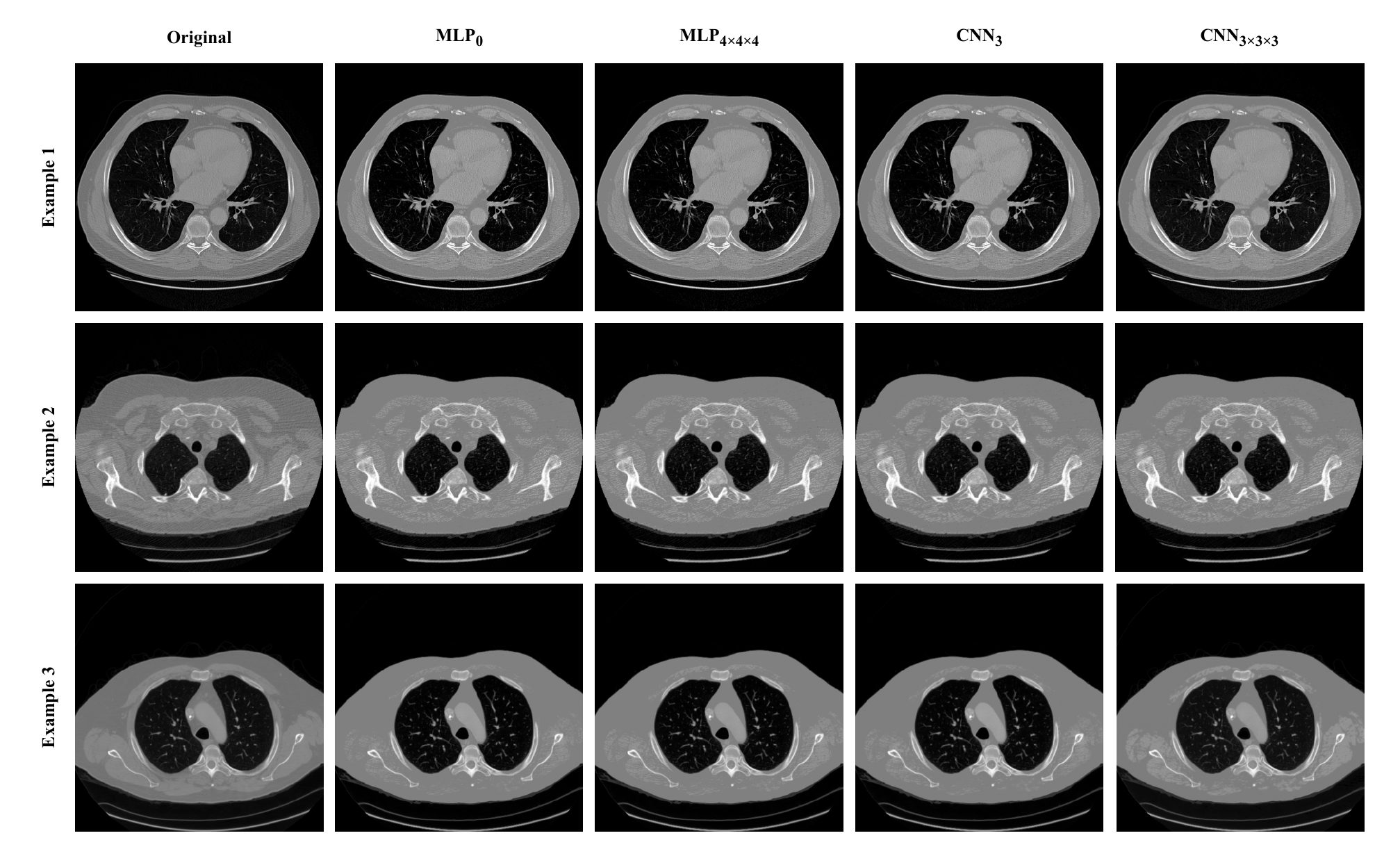}
        \caption{Three full-range CT images reconstructed from four distinct models. The first column presents the original content of the full-range CT slices. The following columns frame the outputs of the reconstruction models MLP$_0$, MLP$_{4 \times 4 \times 4}$, CNN$_3$, and CNN$_{3 \times 3 \times 3}$, respectively, given the HU-clipped images of the original slice.}
        \label{fig:ct-reconstruction-examples}
    \end{figure}

    The performance of the proposed methods for full-range CT reconstruction from HU windows of those same scans is evaluated using MSE, FID, MMD, Precision, Recall and MS-SSIM. The quantitative results are summarized in Table \ref{tab:results-ct-reconstruction}, where bold-accentuated values indicate the best value obtained for the given performance metric. Additionally, visual examples of reconstructed test set slices from different MLP and CNN models are shown in Figure \ref{fig:ct-reconstruction-examples}.
    
    \begin{table}[tb]
        \centering
        \caption{MSE, FID, MMD, Precision, Recall and MS-SSIM metrics for each full-rage CT reconstruction model.}
        {\scriptsize \begin{NiceTabular}{lcccccc}
\hline
\textbf{\begin{tabular}[c]{@{}l@{}}Reconstruction\\ Model\end{tabular}} &
  \textbf{MSE $(\times 10^{-4}) \downarrow$} &
  \textbf{FID $\downarrow$} &
  \textbf{MMD $(\times 10^{-2}) \downarrow$} &
  \textbf{Precision $\uparrow$} &
  \textbf{Recall $\uparrow$} &
  \textbf{MS-SSIM $\uparrow$} \\ \hline
  MLP$_0$                     & 4.68 & 44.9 & 3.29 & 0.969 & 0.989 & 0.962 \\ \hline
  MLP$_4$                     & 4.74 & 44.6 & 3.21 & 0.974 & 0.988 & 0.962 \\ \hline
  MLP$_{4 \times 4}$          & 4.74 & 44.6 & 3.21 & 0.974 & 0.988 & 0.962 \\ \hline
  MLP$_{4 \times 4 \times 4}$ & 4.46 & 41.9 & 2.93 & 0.977 & 0.995 & 0.964 \\ \hline
  CNN$_3$                     & 4.62 & 40.4 & 2.66 & 0.989 & 0.993 & 0.962 \\ \hline
  CNN$_7$                     & 4.54 & 40.1 & 2.68 & 0.988 & 0.993 & 0.962 \\ \hline
  CNN$_11$                    & 4.53 & 40.7 & 2.79 & 0.987 & 0.991 & 0.962 \\ \hline
  CNN$_{3 \times 3}$          & 3.45 & 34.8 & 2.15 & 0.995 & 0.996 & 0.979 \\ \hline
  CNN$_{3 \times 3 \times 3}$ & \bf{2.49} & \bf{26.0} & \bf{1.56} & \bf{0.999} & \bf{1.000} & \bf{0.985} \\ \hline
\end{NiceTabular}}
        \label{tab:results-ct-reconstruction}
    \end{table}

    It is verified that variations of the MLP architecture show similar results across all metrics, with MSE values ranging from 4.46 $\times 10^{-4}$ to 4.74 $\times 10^{-4}$, FID values between 41.9 and 44.9, and MMD values from 2.93 $\times 10^{-2}$ to 3.29 $\times 10^{-2}$, while the CNN variants with increasing convolutional kernel size and depth display greater incremental improvements. Among all configurations, CNN$_{3 \times 3 \times 3}$ obtained the best overall performance, reporting the lowest MSE of 2.49 $\times 10^{-4}$, FID of 26.0, and MMD of 1.56 $\times 10^{-2}$, while also achieving the best Precision and Recall values of 0.999 and 1.000 respectively, and the highest structural similarity with an MS-SSIM of 0.985.

    Moreover, reconstructed images from convolutional models exhibit smoother transitions between tissues of different HU intensities, although the images produced by MLP-based reconstruction models are identically detailed, where subtle differences are noticeable in some cherry-picked / specific regions. These qualitative differences between MLP and CNN reconstruction models can be attributed to their distinct inductive biases. While MLP architectures process voxels independently and therefore tend to preserve sharper local intensity transitions, convolutional models employ weight sharing and local receptive fields, which promote spatial coherence and smoother tissue boundaries, similar to what is present in real medical images.

    \subsection{CT and HU Interval Generation}
    Next, the assessment of the HU generative models is executed not only for the entire HU range with the reconstructed CT image from the HU samples, but also across each HU interval image domain. Such stratified evaluation allows for a finer assessment of each model's ability to reproduce tissue-specific image characteristics. The results are reported separately for each metric in the following sections.

    Apart from Table \ref{tab:results-ms-ssim}, for all the remaining results tables, light green and light red cells indicate better and worse performance compared to the corresponding baseline, respectively, while accentuated shades highlight the best and worst metric values within each generative model type. Furthermore, values in bold point to the approach that achieved the best result across all experiments.

        \subsubsection{Qualitative Examples of Synthetic Lung CT and HU Samples}
\begin{figure}[tb]
            \centering
            \includegraphics[width=1.0\linewidth]{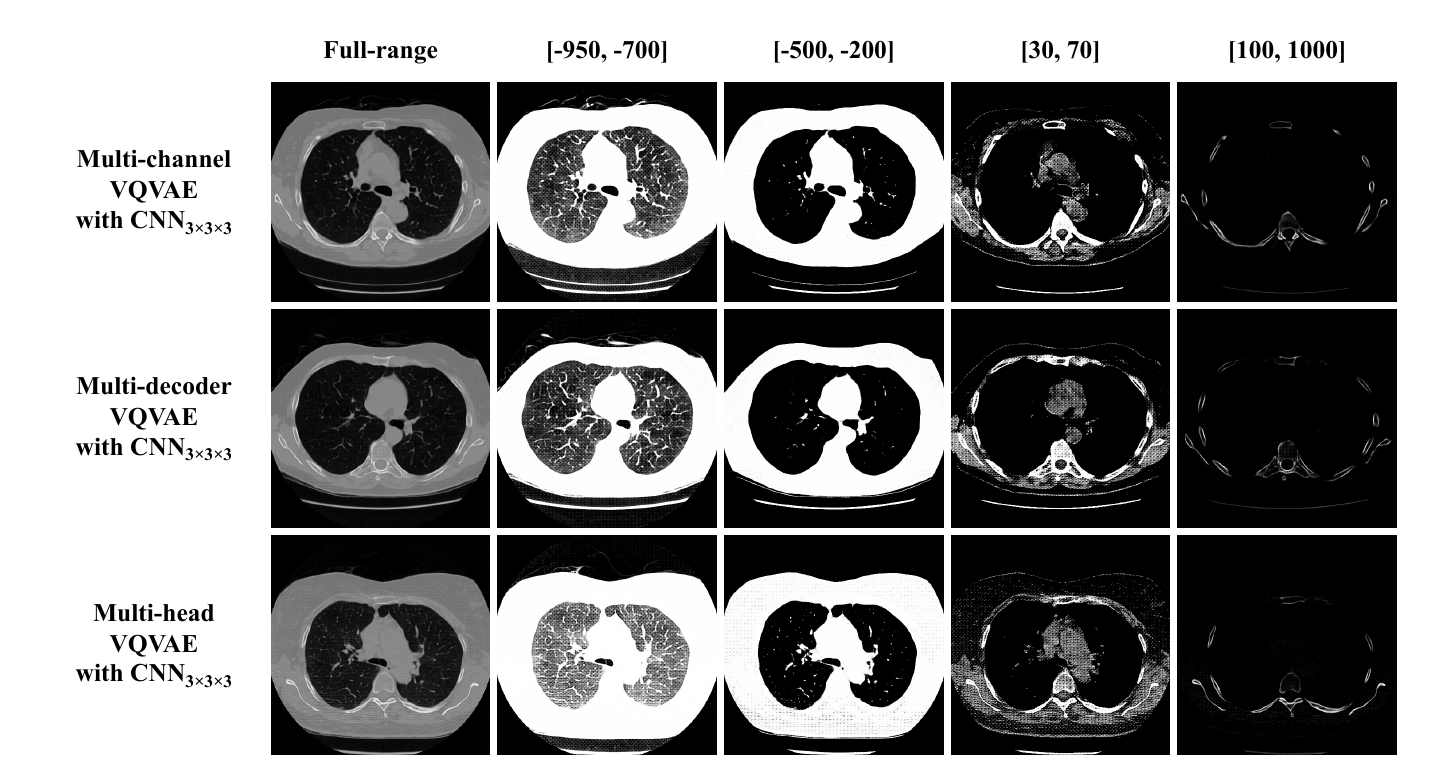}
            \caption{Full-range CT and HU-windowed samples obtained from the proposed VQVAE methods and CNN$_{3 \times 3 \times 3}$ reconstruction model.}
            \label{fig:initial-samples}
        \end{figure}

        This section presents examples of synthetic lung CT images generated by the VQVAE variants of the proposed method to qualitatively provide an illustrative overview of the visual characteristics achievable by the generative framework. Figure \ref{fig:initial-samples} illustrates synthetic samples from the aforementioned generative pipelines.

        The shown examples exhibit anatomically plausible lung structures with spatial coherence across HU windows, since lung parenchyma, surrounding soft tissue and skeleton structures are consistently preserved, and the reconstructed full-range images maintain clear correspondence with the windowed representations. These observations suggest effective learning of the anatomical and textural details for each HU interval domain and ability to produce structurally aligned images during sampling.

        On the other hand, minor grid-like texture patterns can be observed in some regions of the generated images, such as in the samples of the $[30, 70]$ HU interval. These artifacts are commonly caused by the use of perceptual loss terms \citep{johnson2016perceptual}. Importantly, however, such patterns do not alter the overall spatial coherence of the lung structures or alignment across HU windows, primarily affecting only local textural appearance.

        While these qualitative analysis highlight preliminary sample quality, the following sections present comprehensive quantitative results across models and metrics.

        \subsubsection{Assessment by FID}
        Firstly, Table \ref{tab:results-fid} reports the FID values computed per HU interval to assess the visual realism of the generated CT samples from all experiments.
        
        \begin{table}[tb]
            \centering
            \caption{FID computed per HU interval for each generative model configuration.}
            {\scriptsize \begin{NiceTabular}{lllccccc}
\hline
\multirow{2}{*}{\textbf{\begin{tabular}[c]{@{}l@{}}Model\\ Type\end{tabular}}} &
  \multirow{2}{*}{\textbf{Approach}} &
  \multirow{2}{*}{\textbf{\begin{tabular}[c]{@{}l@{}}Reconstruction\\ Model\end{tabular}}} &
  \multicolumn{5}{c}{\textbf{FID per HU Interval}} \\ \cline{4-8} 
 &
   &
   &
  \textbf{Full-range} &
  $\mathbf{[-950, -700]}$ &
  $\mathbf{[-500, -200]}$ &
  $\mathbf{[30, 70]}$ &
  $\mathbf{[100, 1000]}$ \\ \hline
\multirow{3}{*}{WGAN-GP} &
  Baseline &
  - &
  141.8 &
  117.5 &
  131.5 &
  157.2 &
  177.5 \\ \cline{2-8} 
 &
  \multirow{2}{*}{Multi-channel} &
  MLP$_0$ &
  \cc{myred}{211.3} &
  \cc{myred}{196.1} &
  \cc{myred}{137.7} &
  \cc{mygreen}{134.7} &
  \cc{myred}{219.4} \\ \cline{3-8} 
 &
   &
  CNN$_{3 \times 3 \times 3}$ &
  \cc{mylightred}{184.9} &
  \cc{mylightred}{145.0} &
  \cc{mygreen}{121.9} &
  \cc{mylightgreen}{138.2} &
  \cc{mylightred}{205.9} \\ \hline
\multirow{3}{*}{\begin{tabular}[c]{@{}l@{}}Score-based\\ DM\end{tabular}} &
  Baseline &
  - &
  66.8 &
  81.3 &
  56.3 &
  85.6 &
  81.2 \\ \cline{2-8} 
 &
  \multirow{2}{*}{Multi-channel} &
  MLP$_0$ &
  \cc{mylightred}{84.9} &
  \cc{mylightred}{106.5} &
  \cc{mylightred}{134.9} &
  \cc{mylightred}{134.5} &
  \cc{mylightred}{192.3} \\ \cline{3-8} 
 &
   &
  CNN$_{3 \times 3 \times 3}$ &
  \cc{myred}{85.9} &
  \cc{myred}{114.0} &
  \cc{myred}{140.4} &
  \cc{myred}{139.7} &
  \cc{myred}{194.7} \\ \hline
\multirow{7}{*}{VQVAE} &
  Baseline &
  - &
  71.6 &
  95.5 &
  65.2 &
  116.6 &
  105.5 \\ \cline{2-8} 
 &
  \multirow{2}{*}{Multi-channel} &
  MLP$_0$ &
  \cc{myred}{76.3} &
  \cc{mygreen}{\textbf{75.8}} &
  \cc{mygreen}{\textbf{52.9}} &
  \cc{mylightgreen}{83.7} &
  \cc{mylightgreen}{74.6} \\ \cline{3-8} 
 &
   &
  CNN$_{3 \times 3 \times 3}$ &
  \cc{mylightred}{75.7} &
  \cc{mylightgreen}{77.6} &
  \cc{mylightred}{66.1} &
  \cc{mylightgreen}{79.2} &
  \cc{mylightgreen}{75.3} \\ \cline{2-8} 
 &
  \multirow{2}{*}{Multi-decoder} &
  MLP$_0$ &
  \cc{mylightgreen}{70.5} &
  \cc{myred}{100.0} &
  \cc{myred}{70.5} &
  \cc{mylightgreen}{86.7} &
  \cc{mylightgreen}{76.3} \\ \cline{3-8} 
 &
   &
  CNN$_{3 \times 3 \times 3}$ &
  \cc{mylightgreen}{67.9} &
  \cc{mylightgreen}{81.4} &
  \cc{mylightred}{69.2} &
  \cc{mylightgreen}{77.4} &
  \cc{mylightgreen}{77.2} \\ \cline{2-8} 
 &
  \multirow{2}{*}{Multi-head} &
  MLP$_0$ &
  \cc{mylightgreen}{68.1} &
  \cc{mylightred}{96.3} &
  \cc{mylightgreen}{63.6} &
  \cc{mylightgreen}{73.7} &
  \cc{mygreen}{\textbf{71.3}} \\ \cline{3-8} 
 &
   &
  CNN$_{3 \times 3 \times 3}$ &
  \cc{mygreen}{\textbf{67.1}} &
  \cc{mylightgreen}{77.0} &
  \cc{mylightgreen}{59.4} &
  \cc{mygreen}{\textbf{73.1}} &
  \cc{mylightgreen}{71.5} \\ \hline
\end{NiceTabular}}
            \includegraphics[width=0.75\linewidth]{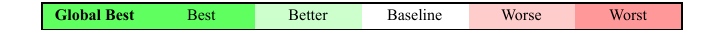}
            \label{tab:results-fid}
        \end{table}

        The baseline configurations highlight the performance difference across the generative model families, where the score-based DM and VQVAE models achieved considerably lower FID values (66.8 and 71.6 in full-range reconstruction, respectively) than the WGAN-GP's (141.8). Moreover, the score-based DM obtained the lowest FID values for all HU intervals used in the work, ranging from 56.3 to 85.6 across the tested ranges.
        
        When analysing the multi-channel configurations, the results demonstrate that the approach improves the generative fidelity of HU windows images in the VQVAEs experiments, with the MLP$_0$ configuration achieving values as low as 52.9 in the $[-500, -200]$ interval, while maintaining comparable performance to the baseline in the full-range reconstructed domain (76.3 and 75.7 for MLP$_0$ and CNN$_{3 \times 3 \times 3}$ variants respectively, compared to the baseline's 71.6). In contrast, for WGAN-GP and DM models, the multi-channel configuration did not provide consistent gains, and in some cases led to performance degradation, with WGAN-GP reaching 211.3 and 184.9 in full-range reconstruction for MLP$_0$ and CNN$_{3 \times 3 \times 3}$ reconstruction network configurations respectively. The multi-decoder VQVAE configurations yielded noticeable improvements in the $[30, 70]$ and $[100, 1000]$ HU ranges, achieving 77.4 and 77.2 respectively with the CNN-based reconstruction model, while slightly under-performing in the remaining intervals. Despite not achieving uniformly lower FID values across all HU domains, the multi-decoder approach maintains baseline-level performance in full-range reconstruction (70.5 and 67.9 for MLP$_0$ and CNN$_{3 \times 3 \times 3}$ variants) while providing HU-specific samples whose feature distributions more closely match those of the original data. The multi-head configurations follow a similar trend but demonstrate superior overall performance, achieving the lowest FID values across most HU intervals, including the full-range domain with values of 68.1 and 67.1 using MLP$_0$ and CNN$_{3 \times 3 \times 3}$ respectively.
        
        It is also noted that the multi-head VQVAE with the CNN$_{3 \times 3 \times 3}$ reconstruction model was the only generative pipeline to obtain better results than the baseline approach in all HU ranges, as well as the experiment with lowest FID globally with a value of 67.1 in full-range reconstruction.
        
        Lastly, when comparing the results between the same generative methods with different CT reconstruction networks, the majority of the approaches employing the convolutional network outperformed the ones using MLP models.

        \subsubsection{Assessment by MMD}
        Afterwards, Table \ref{tab:results-mmd} displays MMD values, comparing the discrepancy between the distributions of real and generated samples across HU intervals, for each generative pipeline configuration.
        
        \begin{table}[tb]
            \centering
            \caption{MMD computed per HU interval for each generative model configuration.}
            {\scriptsize \begin{NiceTabular}{lllccccc}
\hline
\multirow{2}{*}{\textbf{\begin{tabular}[c]{@{}l@{}}Model\\ Type\end{tabular}}} &
  \multirow{2}{*}{\textbf{Approach}} &
  \multirow{2}{*}{\textbf{\begin{tabular}[c]{@{}l@{}}Reconstruction\\ Model\end{tabular}}} &
  \multicolumn{5}{c}{\textbf{MMD ($\times 10^{-2}$) per HU Interval}} \\ \cline{4-8} 
 &
   &
   &
  \textbf{Full-range} &
  $\mathbf{[-950, -700]}$ &
  $\mathbf{[-500, -200]}$ &
  $\mathbf{[30, 70]}$ &
  $\mathbf{[100, 1000]}$ \\ \hline
\multirow{3}{*}{WGAN-GP} &
  Baseline &
  - &
  17.00 &
  8.93 &
  14.51 &
  15.78 &
  16.24 \\ \cline{2-8} 
 &
  \multirow{2}{*}{Multi-channel} &
  MLP$_0$ &
  \cc{myred}{28.06} &
  \cc{myred}{19.96} &
  \cc{myred}{15.08} &
  \cc{mygreen}{9.00} &
  \cc{myred}{20.56} \\ \cline{3-8} 
 &
   &
  CNN$_{3 \times 3 \times 3}$ &
  \cc{mylightred}{23.14} &
  \cc{mylightred}{11.84} &
  \cc{mygreen}{13.19} &
  \cc{mylightgreen}{9.94} &
  \cc{mylightred}{18.69} \\ \hline
\multirow{3}{*}{\begin{tabular}[c]{@{}l@{}}Score-based\\ DM\end{tabular}} &
  Baseline &
  - &
  4.81 &
  5.31 &
  2.69 &
  3.42 &
  4.07 \\ \cline{2-8} 
 &
  \multirow{2}{*}{Multi-channel} &
  MLP$_0$ &
  \cc{myred}{9.84} &
  \cc{myred}{9.69} &
  \cc{myred}{17.72} &
  \cc{mylightred}{12.98} &
  \cc{mylightred}{20.53} \\ \cline{3-8} 
 &
   &
  CNN$_{3 \times 3 \times 3}$ &
  \cc{mylightred}{7.95} &
  \cc{mylightred}{9.55} &
  \cc{mylightred}{17.54} &
  \cc{myred}{13.26} &
  \cc{myred}{20.76} \\ \hline
\multirow{7}{*}{VQVAE} &
  Baseline &
  - &
  6.31 &
  8.91 &
  4.54 &
  7.28 &
  8.24 \\ \cline{2-8} 
 &
  \multirow{2}{*}{Multi-channel} &
  MLP$_0$ &
  \cc{mylightred}{6.56} &
  \cc{mylightgreen}{5.51} &
  \cc{mygreen}{\textbf{2.92}} &
  \cc{mylightgreen}{3.84} &
  \cc{mygreen}{\textbf{2.03}} \\ \cline{3-8} 
 &
   &
  CNN$_{3 \times 3 \times 3}$ &
  \cc{myred}{6.76} &
  \cc{mylightgreen}{4.90} &
  \cc{mylightgreen}{4.27} &
  \cc{mylightgreen}{3.35} &
  \cc{mylightgreen}{3.54} \\ \cline{2-8} 
 &
  \multirow{2}{*}{Multi-decoder} &
  MLP$_0$ &
  \cc{mylightgreen}{5.50} &
  \cc{mylightgreen}{6.84} &
  \cc{myred}{4.93} &
  \cc{mylightgreen}{3.97} &
  \cc{mylightgreen}{2.14} \\ \cline{3-8} 
 &
   &
  CNN$_{3 \times 3 \times 3}$ &
  \cc{mylightgreen}{5.20} &
  \cc{mylightgreen}{4.89} &
  \cc{mylightred}{4.85} &
  \cc{mylightgreen}{2.97} &
  \cc{mylightgreen}{2.47} \\ \cline{2-8} 
 &
  \multirow{2}{*}{Multi-head} &
  MLP$_0$ &
  \cc{mygreen}{\textbf{4.62}} &
  \cc{mylightgreen}{6.56} &
  \cc{mylightred}{4.79} &
  \cc{mylightgreen}{2.83} &
  \cc{mylightgreen}{2.12} \\ \cline{3-8} 
 &
   &
  CNN$_{3 \times 3 \times 3}$ &
  \cc{mylightgreen}{5.17} &
  \cc{mygreen}{\textbf{4.61}} &
  \cc{mylightgreen}{3.07} &
  \cc{mygreen}{\textbf{2.21}} &
  \cc{mylightgreen}{2.09} \\ \hline
\end{NiceTabular}}
            \includegraphics[width=0.75\linewidth]{figures//results/results-scale-horizontal-outer-border.pdf}
            \label{tab:results-mmd}
        \end{table}

        The baseline configurations reveal substantial performance differences across generative model architectures, with the score-based DM achieving the lowest MMD values of 4.81 $\times 10^{-2}$ in full-range reconstruction and ranging from 2.69 $\times 10^{-2}$ to 5.31 $\times 10^{-2}$ across all HU intervals, followed by VQVAE (6.31 $\times 10^{-2}$ in full-range) and WGAN-GP (17.00 $\times 10^{-2}$), similarly to what was registered in the respective FID scores.

        Examining the multi-channel experiments, the results indicate that this approach yielded improvements for VQVAE models across most HU intervals, achieving values as low as 2.92 $\times 10^{-2}$ and 2.03 $\times 10^{-2}$ for the MLP$_0$ configuration in the $[-500, -200]$ and $[100, 1000]$ intervals respectively, while maintaining competitive performance in full-range reconstruction with 6.56 $\times 10^{-2}$ compared to the baseline's 6.31 $\times 10^{-2}$. For WGAN-GP models, the multi-channel configuration showed mostly negative results, with full-range values increasing to 28.06 $\times 10^{-2}$ and 23.14 $\times 10^{-2}$, although some improvements were observed in certain HU intervals such as $[-500, -200]$ and $[30, 70]$. The score-based DM multi-channel pipeline exhibited worse performance for all HU ranges compared to its baseline model, with full-range values of 9.84 $\times 10^{-2}$ and 7.95 $\times 10^{-2}$ for MLP$_0$ and CNN$_{3 \times 3 \times 3}$ variants respectively. On the other hand, the multi-decoder VQVAE configurations demonstrated significant improvements for all HU ranges, achieving 4.89 $\times 10^{-2}$, 2.97 $\times 10^{-2}$, and 2.47 $\times 10^{-2}$ for the $[-950, -700]$, $[30, 70]$, and $[100, 1000]$ intervals respectively with the MLP$_0$ model, except in $[-500, -200]$ where it displayed lower but close MMD values of 4.85 $\times 10^{-2}$. However, the multi-head VQVAE experiments demonstrated consistent increased performance for all HU domains in the CNN$_{3 \times 3 \times 3}$ variant, with full-range value of 5.17 $\times 10^{-2}$, while also detaining the lowest MMD values of 4.61 $\times 10^{-2}$ and 2.21 $\times 10^{-2}$ for the $[-950, -700]$ and $[30, 70]$ intervals respectively. Even so, the multi-head variant employing the MLP$_0$ reconstruction network obtained the lowest full-range MMD of all experiments with value 4.62 $\times 10^{-2}$.

        Comparing experiments that use different reconstruction networks, the convolutional models yield better results than their MLP counterparts across most configurations and HU intervals.

        \subsubsection{Assessment by Precision}
        Table \ref{tab:results-precision} frames the Precision values obtained for each experiment, reflecting each model's fidelity of generated samples relative to the real data distribution.
        
        \begin{table}[tb]
            \centering
            \caption{Precision computed per HU interval for each generative model configuration.}
            {\scriptsize \begin{NiceTabular}{lllccccc}
\hline
\multirow{2}{*}{\textbf{\begin{tabular}[c]{@{}l@{}}Model\\ Type\end{tabular}}} &
  \multirow{2}{*}{\textbf{Approach}} &
  \multirow{2}{*}{\textbf{\begin{tabular}[c]{@{}l@{}}Reconstruction\\ Model\end{tabular}}} &
  \multicolumn{5}{c}{\textbf{Precision per HU Interval}} \\ \cline{4-8} 
 &
   &
   &
  \textbf{Full-range} &
  $\mathbf{[-950, -700]}$ &
  $\mathbf{[-500, -200]}$ &
  $\mathbf{[30, 70]}$ &
  $\mathbf{[100, 1000]}$ \\ \hline
\multirow{3}{*}{WGAN-GP} &
  Baseline &
  - &
  0.022 &
  0.121 &
  0.132 &
  0.262 &
  0.056 \\ \cline{2-8} 
 &
  \multirow{2}{*}{Multi-channel} &
  MLP$_0$ &
  \cc{myred}{0.001} &
  \cc{myred}{0.004} &
  \cc{myred}{0.122} &
  \cc{mygreen}{0.269} &
  \cc{mylightred}{0.034} \\ \cline{3-8} 
 &
   &
  CNN$_{3 \times 3 \times 3}$ &
  \cc{myred}{0.001} &
  \cc{mylightred}{0.062} &
  \cc{mygreen}{0.213} &
  \cc{myred}{0.210} &
  \cc{myred}{0.029} \\ \hline
\multirow{3}{*}{\begin{tabular}[c]{@{}l@{}}Score-based\\ DM\end{tabular}} &
  Baseline &
  - &
  \textbf{0.747} &
  0.642 &
  0.780 &
  0.680 &
  \textbf{0.739} \\ \cline{2-8} 
 &
  \multirow{2}{*}{Multi-channel} &
  MLP$_0$ &
  \cc{mylightred}{0.311} &
  \cc{mylightred}{0.218} &
  \cc{myred}{0.066} &
  \cc{mylightred}{0.313} &
  \cc{myred}{0.182} \\ \cline{3-8} 
 &
   &
  CNN$_{3 \times 3 \times 3}$ &
  \cc{myred}{0.208} &
  \cc{myred}{0.208} &
  \cc{mylightred}{0.068} &
  \cc{myred}{0.234} &
  \cc{mylightred}{0.183} \\ \hline
\multirow{7}{*}{VQVAE} &
  Baseline &
  - &
  0.639 &
  0.452 &
  0.634 &
  0.512 &
  0.719 \\ \cline{2-8} 
 &
  \multirow{2}{*}{Multi-channel} &
  MLP$_0$ &
  \cc{myred}{0.382} &
  \cc{mygreen}{\textbf{0.653}} &
  \cc{mygreen}{\textbf{0.789}} &
  \cc{mylightgreen}{0.666} &
  \cc{myred}{0.648} \\ \cline{3-8} 
 &
   &
  CNN$_{3 \times 3 \times 3}$ &
  \cc{mylightred}{0.416} &
  \cc{mylightgreen}{0.629} &
  \cc{mylightred}{0.608} &
  \cc{mylightgreen}{0.652} &
  \cc{mylightred}{0.706} \\ \cline{2-8} 
 &
  \multirow{2}{*}{Multi-decoder} &
  MLP$_0$ &
  \cc{mylightred}{0.448} &
  \cc{myred}{0.348} &
  \cc{myred}{0.559} &
  \cc{mylightgreen}{0.545} &
  \cc{mylightred}{0.689} \\ \cline{3-8} 
 &
   &
  CNN$_{3 \times 3 \times 3}$ &
  \cc{mylightred}{0.564} &
  \cc{mylightgreen}{0.541} &
  \cc{mylightred}{0.584} &
  \cc{mylightgreen}{0.651} &
  \cc{mylightred}{0.652} \\ \cline{2-8} 
 &
  \multirow{2}{*}{Multi-head} &
  MLP$_0$ &
  \cc{mylightred}{0.454} &
  \cc{myred}{0.348} &
  \cc{mylightgreen}{0.700} &
  \cc{mygreen}{\textbf{0.724}} &
  \cc{mylightred}{0.713} \\ \cline{3-8} 
 &
   &
  CNN$_{3 \times 3 \times 3}$ &
  \cc{mylightred}{0.576} &
  \cc{mylightgreen}{0.617} &
  \cc{mylightgreen}{0.702} &
  \cc{mylightgreen}{0.705} &
  \cc{mylightred}{0.708} \\ \hline
\end{NiceTabular}}
            \includegraphics[width=0.75\linewidth]{figures//results/results-scale-horizontal-outer-border.pdf}
            \label{tab:results-precision}
        \end{table}

        The baseline experiments demonstrate considerable variation in Precision across generative model families, with the score-based DM achieving the highest values in most HU intervals, followed by VQVAE and WGAN-GP. The score-based DM baseline obtained Precision values over 0.642 across all tested intervals, indicating strong alignment of generated samples with the original data manifold.

        When observing the multi-channel configurations, the results show that his approach failed to improve Precision values for WGAN-GP and score-based DM models compared to their respective baselines across most HU intervals. For WGAN-GP, the multi-channel configuration with the convolutional reconstruction model showed great degradation in full-range Precision (0.001) compared to the baseline (0.022), although improvements were observed in specific intervals such as $[-950, -700]$ and $[-500, -200]$. The score-based DM multi-channel approach similarly showed reduced Precision in the complete HU range, declining from 0.747 to 0.208 for the configuration paired with the convolutional reconstruction network.

        The VQVAE multi-channel, multi-decoder, and multi-head experiments all demonstrated reduced Precision values in full-range reconstruction relative to the baseline, with values ranging from 0.382 to 0.576 compared to the baseline value of 0.639. However, these methods achieved Precision values in specific HU intervals that exceed, or in certain cases closely approximate, those of the VQVAE baseline experiment. Notably, the multi-channel configuration with an MLP reconstruction network achieved 0.653 in the $[-950, -700]$ interval and 0.789 in the $[-500, -200]$ interval, the highest Precision values obtained across all experiments. The multi-head configuration together with CNN$_{3 \times 3 \times 3}$ demonstrated the most consistent Precision performance across all HU domains, with all values ranging from 0.576 to 0.708.

        \subsubsection{Assessment by Recall}
        Recall is used to assess the diversity of generated samples, indicating how well each model covers the original data distribution. Table \ref{tab:results-recall} lists Recall scores per HU interval for each generative experiment.
        
        \begin{table}[tb]
            \centering
            \caption{Recall computed per HU interval for each generative model configuration.}
            {\scriptsize \begin{NiceTabular}{lllccccc}
\hline
\multirow{2}{*}{\textbf{\begin{tabular}[c]{@{}l@{}}Model\\ Type\end{tabular}}} &
  \multirow{2}{*}{\textbf{Approach}} &
  \multirow{2}{*}{\textbf{\begin{tabular}[c]{@{}l@{}}Reconstruction\\ Model\end{tabular}}} &
  \multicolumn{5}{c}{\textbf{Recall per HU Interval}} \\ \cline{4-8} 
 &
   &
   &
  \textbf{Full-range} &
  $\mathbf{[-950, -700]}$ &
  $\mathbf{[-500, -200]}$ &
  $\mathbf{[30, 70]}$ &
  $\mathbf{[100, 1000]}$ \\ \hline
\multirow{3}{*}{WGAN-GP} &
  Baseline &
  - &
  0.016 &
  0.060 &
  0.048 &
  0.023 &
  0.006 \\ \cline{2-8} 
 &
  \multirow{2}{*}{Multi-channel} &
  MLP$_0$ &
  \cc{mylightred}{0.003} &
  \cc{myred}{0.005} &
  \cc{myred}{0.045} &
  \cc{mygreen}{0.079} &
  \cc{myred}{0.002} \\ \cline{3-8} 
 &
   &
  CNN$_{3 \times 3 \times 3}$ &
  \cc{myred}{0.000} &
  \cc{mylightred}{0.038} &
  \cc{mygreen}{0.119} &
  \cc{mylightgreen}{0.078} &
  \cc{mygreen}{0.008} \\ \hline
\multirow{3}{*}{\begin{tabular}[c]{@{}l@{}}Score-based\\ DM\end{tabular}} &
  Baseline &
  - &
  0.263 &
  0.282 &
  0.483 &
  0.427 &
  0.464 \\ \cline{2-8} 
 &
  \multirow{2}{*}{Multi-channel} &
  MLP$_0$ &
  \cc{myred}{0.100} &
  \cc{mylightred}{0.073} &
  \cc{mylightred}{0.027} &
  \cc{mylightred}{0.044} &
  \cc{mylightred}{0.007} \\ \cline{3-8} 
 &
   &
  CNN$_{3 \times 3 \times 3}$ &
  \cc{mylightred}{0.127} &
  \cc{myred}{0.060} &
  \cc{myred}{0.024} &
  \cc{myred}{0.034} &
  \cc{myred}{0.004} \\ \hline
\multirow{7}{*}{VQVAE} &
  Baseline &
  - &
  0.179 &
  0.109 &
  0.407 &
  0.235 &
  0.301 \\ \cline{2-8} 
 &
  \multirow{2}{*}{Multi-channel} &
  MLP$_0$ &
  \cc{myred}{0.160} &
  \cc{mygreen}{\textbf{0.263}} &
  \cc{mygreen}{\textbf{0.470}} &
  \cc{mylightgreen}{0.307} &
  \cc{mylightgreen}{0.509} \\ \cline{3-8} 
 &
   &
  CNN$_{3 \times 3 \times 3}$ &
  \cc{mylightgreen}{0.205} &
  \cc{mylightgreen}{0.202} &
  \cc{mylightgreen}{0.435} &
  \cc{mylightgreen}{0.377} &
  \cc{mylightgreen}{0.507} \\ \cline{2-8} 
 &
  \multirow{2}{*}{Multi-decoder} &
  MLP$_0$ &
  \cc{mygreen}{\textbf{0.354}} &
  \cc{mylightgreen}{0.188} &
  \cc{mylightred}{0.366} &
  \cc{mylightgreen}{0.334} &
  \cc{mygreen}{\textbf{0.538}} \\ \cline{3-8} 
 &
   &
  CNN$_{3 \times 3 \times 3}$ &
  \cc{mylightgreen}{0.217} &
  \cc{mylightgreen}{0.202} &
  \cc{mylightred}{0.382} &
  \cc{mygreen}{\textbf{0.459}} &
  \cc{mylightgreen}{0.486} \\ \cline{2-8} 
 &
  \multirow{2}{*}{Multi-head} &
  MLP$_0$ &
  \cc{mylightgreen}{0.334} &
  \cc{mylightgreen}{0.243} &
  \cc{myred}{0.335} &
  \cc{mylightgreen}{0.424} &
  \cc{mylightgreen}{0.474} \\ \cline{3-8} 
 &
   &
  CNN$_{3 \times 3 \times 3}$ &
  \cc{mylightgreen}{0.266} &
  \cc{mylightgreen}{0.239} &
  \cc{mylightgreen}{0.436} &
  \cc{mylightgreen}{0.432} &
  \cc{mylightgreen}{0.482} \\ \hline
\end{NiceTabular}}
            \includegraphics[width=0.75\linewidth]{figures//results/results-scale-horizontal-outer-border.pdf}
            \label{tab:results-recall}
        \end{table}

        The results from the baseline methods show that there is some difference between Recall performances across generative model types, with the score-based DM baseline achieving the highest values across all HU intervals, ranging from 0.263 to 0.483. The VQVAE baseline obtained Recall values between 0.109 and 0.407, while WGAN-GP demonstrated the lowest performance with values no exceeding 0.060 in any interval.

        When examining the multi-channel configurations, the results indicate that this approach did not surpass baseline performance for WGAN-GP and score-based DM models in most HU intervals. Despite showing some improvements in some HU intervals, the WGAN-GP multi-channel configurations showed very poor Recall values across all HU domains. The score-based DM multi-channel approaches displayed worse performance for all HU ranges, with the pipeline using the MLP reconstruction model achieving a Recall of 0.100 on full-range CT samples, and the convolutional version yielding 0.127.

        In contrast, the VQVAE multi-channel, multi-decoder, and multi-head configurations displayed substantial improvements in Recall values compared to the baseline across both full-range and HU-restricted domains. The multi-channel approach with MLP reconstruction network outputted the overall best Recall values for the $[-950, -700]$ and $[-500, -200]$ intervals, albeit achieving the worst VQVAE Recall value for full-range CT samples, similar to the baseline's value. The multi-decoder approach coupled with MLP$_0$ obtained the highest full-range Recall among all experiments with 0.354, nearly doubling the baseline performance of 0.179. Furthermore, the multi-decoder approaches also achieved the best global Recall results for the $[30, 70]$ and $[100, 1000]$ ranges. Finally, the multi-head configurations maintained elevated Recall across all intervals, namely the one employing the CNN$_{3 \times 3 \times 3}$ reconstruction model which was one of the experiments that outperformed the baseline across all HU ranges, alongside the multi-channel variant also paired with the convolutional reconstruction network.

        \subsubsection{Assessment of Sample Variety via MS-SSIM}
\begin{figure}[tb]
            \centering
            \includegraphics[width=1.0\linewidth]{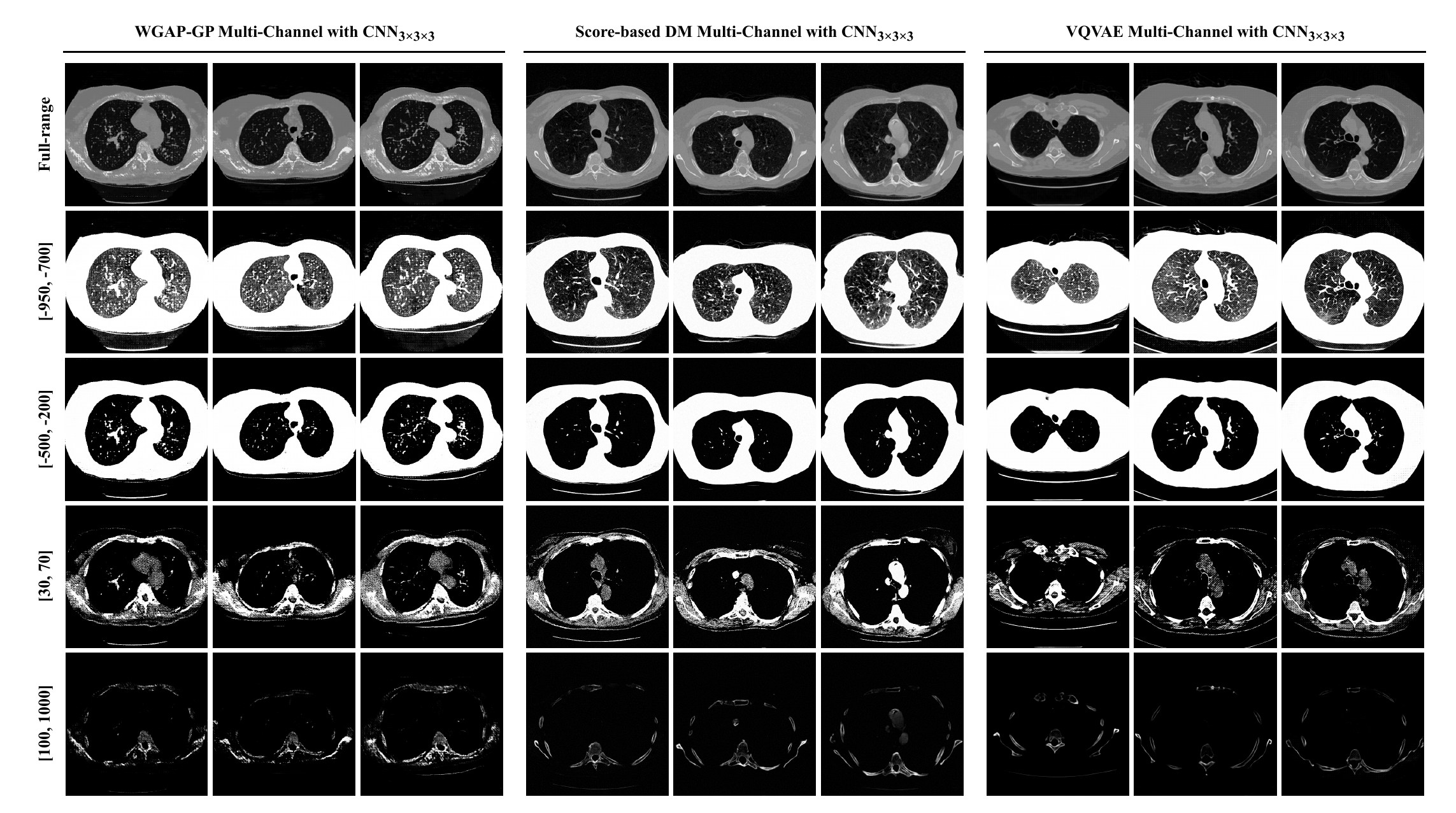}
            \caption{Sample examples with respective full-range and HU-windowed representations from the multi-channel approaches and CNN$_{3 \times 3 \times 3}$ reconstruction model.}
            \label{fig:ms-ssim-sample-examples}
        \end{figure}

        Finally, the intra-sample variety of generated images using MS-SSIM is examined. Table \ref{tab:results-ms-ssim} shows MS-SSIM values across HU intervals, where lower scores indicate higher diversity among generated samples. For each generative model type and HU range, the lowest MS-SSIM value is underlined, and the global lowest values for each HU range are accentuated in bold. Moreover, some samples of different generative approaches are pictured in Figure \ref{fig:ms-ssim-sample-examples}, displaying some of the sample variety produced by the models.
        
        \begin{table}[tb]
            \centering
            \caption{MS-SSIM computed per HU interval, over all possible pairs of images of the synthetic sample set, for each generative model configuration.}
            {\scriptsize \begin{NiceTabular}{lllccccc}
\hline
\multirow{2}{*}{\textbf{\begin{tabular}[c]{@{}l@{}}Model\\ Type\end{tabular}}} &
  \multirow{2}{*}{\textbf{Approach}} &
  \multirow{2}{*}{\textbf{\begin{tabular}[c]{@{}l@{}}Reconstruction\\ Model\end{tabular}}} &
  \multicolumn{5}{c}{\textbf{MS-SSIM per HU Interval}} \\ \cline{4-8} 
 &
   &
   &
  \textbf{Full-range} &
  $\mathbf{[-950, -700]}$ &
  $\mathbf{[-500, -200]}$ &
  $\mathbf{[30, 70]}$ &
  $\mathbf{[100, 1000]}$ \\ \hline
\multirow{3}{*}{WGAN-GP} &
  Baseline &
  - &
  0.392 &
  0.371 &
  0.384 &
  \uline{0.371} &
  \uline{0.617} \\ \cline{2-8} 
 &
  \multirow{2}{*}{Multi-channel} &
  MLP$_0$ &
  0.382 &
  \uline{0.335} &
  \uline{0.357} &
  0.402 &
  0.642 \\ \cline{3-8} 
 &
   &
  CNN$_{3 \times 3 \times 3}$ &
  \uline{0.373} &
  0.357 &
  0.373 &
  0.404 &
  0.653 \\ \hline
\multirow{3}{*}{\begin{tabular}[c]{@{}l@{}}Score-based\\ DM\end{tabular}} &
  Baseline &
  - &
  0.343 &
  \uline{0.291} &
  0.356 &
  \uline{0.339} &
  0.659 \\ \cline{2-8} 
 &
  \multirow{2}{*}{Multi-channel} &
  MLP$_0$ &
  0.352 &
  0.293 &
  \uline{0.355} &
  0.340 &
  \uline{0.642} \\ \cline{3-8} 
 &
   &
  CNN$_{3 \times 3 \times 3}$ &
  \uline{0.331} &
  0.296 &
  0.358 &
  0.344 &
  \uline{0.642} \\ \hline
\multirow{7}{*}{VQVAE} &
  Baseline &
  - &
  0.395 &
  0.370 &
  0.405 &
  0.432 &
  0.756 \\ \cline{2-8} 
 &
  \multirow{2}{*}{Multi-channel} &
  MLP$_0$ &
  0.359 &
  0.336 &
  0.363 &
  0.337 &
  \uline{0.646} \\ \cline{3-8} 
 &
   &
  CNN$_{3 \times 3 \times 3}$ &
  0.365 &
  0.329 &
  0.368 &
  0.371 &
  0.754 \\ \cline{2-8} 
 &
  \multirow{2}{*}{Multi-decoder} &
  MLP$_0$ &
  \uline{0.339} &
  \uline{0.309} &
  \uline{0.343} &
  \uline{0.311} &
  0.694 \\ \cline{3-8} 
 &
   &
  CNN$_{3 \times 3 \times 3}$ &
  0.365 &
  0.341 &
  0.368 &
  0.354 &
  0.701 \\ \cline{2-8} 
 &
  \multirow{2}{*}{Multi-head} &
  MLP$_0$ &
  0.365 &
  0.311 &
  0.373 &
  0.363 &
  0.698 \\ \cline{3-8} 
 &
   &
  CNN$_{3 \times 3 \times 3}$ &
  0.362 &
  0.329 &
  0.361 &
  0.354 &
  0.714 \\ \hline
\multicolumn{3}{l}{\textbf{Test Dataset}} &
  \textbf{0.337} &
  \textbf{0.321} &
  \textbf{0.348} &
  \textbf{0.327} &
  \textbf{0.644} \\ \hline
\end{NiceTabular}
}
            \label{tab:results-ms-ssim}
        \end{table}

        Firstly, it is noted that the MS-SSIM was computed for the test set to obtain baseline sample variety scores for all HU ranges. The test set baseline achieved MS-SSIM values of 0.337 for full-range, 0.321 for the $[-950, -700]$ interval, 0.348 for the $[-500, -200]$ interval, 0.327 for the $[30, 70]$ interval, and, lastly, 0.644 for the $[100, 1000]$ interval.

        Considering the baseline experiments, MS-SSIM values range from 0.343 to 0.395, with the score-based DM baseline achieving the lowest value, followed by WGAN-GP and VQVAE. Looking into the multi-channel configurations for WGAN-GP and score-based DM models, reductions in MS-SSIM were observed compared to their respective baselines, although the differences remained small. The WGAN-GP multi-channel experiments achieved MS-SSIM values of 0.382 and 0.373 for the generative pipelines using the MLP$_0$ and CNN$_{3 \times 3 \times 3}$ respectively, while the score-based DM multi-channel approaches obtained 0.352 and 0.331. On the other hand, all VQVAE multi-channel, multi-decoder and multi-head configurations demonstrated lower MS-SSIM values than the baseline across most HU intervals. The multi-decoder configuration with MLP$_0$ yielded the lowest MS-SSIM values among VQVAE methods for almost all HU ranges, yielding a MS-SSIM value of 0.339 for the full-range CT samples.

        Specially, it is important to note that the MS-SSIM values obtained across all experimental configurations remained close to those of the test dataset, with full-range values spanning a relatively narrow range from 0.331 to 0.395. This pattern was also noted across all HU interval results, which further confirms the ability of the generative models to generate common tissue structures found in the images of each HU domain. The variation of MS-SSIM values across HU intervals aligns with known anatomical properties. For example, lung-tissue ranges exhibit lower MS-SSIM values around 0.33, reflecting higher structural variability across samples, while the $[100, -1000]$ interval presents a significantly higher value of approximately 0.64, consistent with the rigid and homogeneous nature of bone structures.

    \subsection{Segmentation Task Qualitative Assessment}
\begin{figure}[tb]
            \centering
            \includegraphics[width=1.0\linewidth]{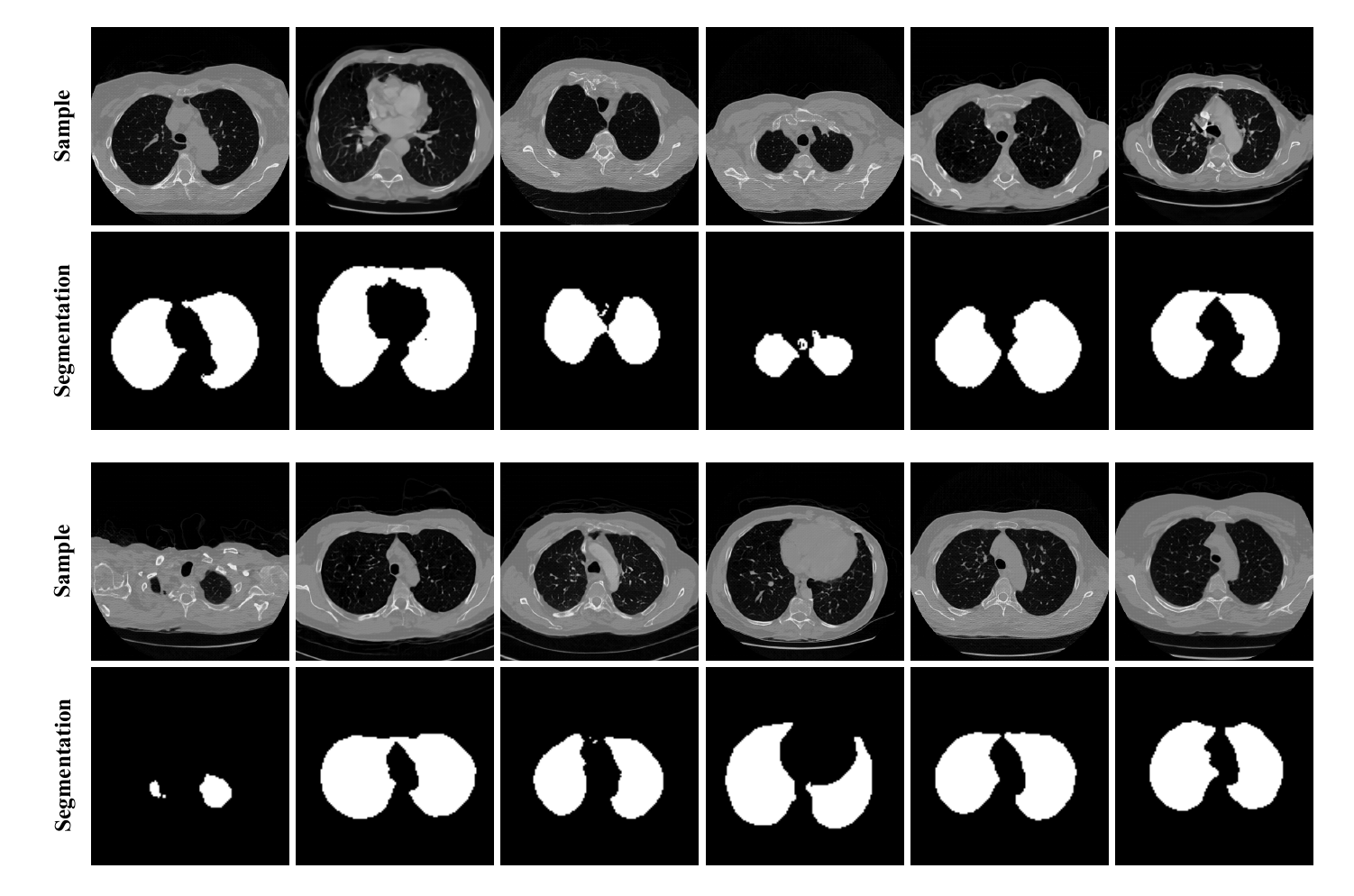}
            \caption{Synthetic full-range CT images sampled from the multi-head VQVAE and CNN$_{3 \times 3 \times 3}$ reconstruction model and respective segmentation masks outputted from the segmentation model from \citep{segmentation-model}.}
            \label{fig:segmentation-results}
        \end{figure}

    Figure \ref{fig:segmentation-results} illustrate generated full-range CT samples produced by the multi-head VQVAE generative model coupled with the CNN$_{3 \times 3 \times 3}$ reconstruction model, as well as their corresponding lung segmentation masks obtained using the segmentation model mentioned in Section \ref{sec:experimental-setup}. Despite the fact that analysis extracted from the Figure can only be qualitative, they provide indirect, yet informative, means of assessing whether the structural content of the synthetic samples is compatible with the expected input distributional properties of the downstream task model.  

    Overall, the segmentation model is able to generate well-defined and anatomically coherent lung masks from the synthetic images, with the predicted segmentation masks consistently capturing the global shape of the lungs, preserving the separation of the left and right lobes, and adapting to variations in lung morphology across sampled images. The fact that a network trained on real CT data can successfully process and segment the generated samples indicates that the proposed generative pipeline produces images that lie within the effective input distribution expected by the segmentation model. This observation supports both the fidelity of the generated images and their utility for downstream tasks, as the synthetic data can be meaningfully interpreted by a separately trained model.

    Although Figure \ref{fig:segmentation-results} demonstrates that the images are correctly processed by the segmentation pipeline, which points to good indirect performance from the proposed method of this work, some segmentation imperfections are visible, particularly in regions proximal to the mediastinum and pericardial cavity. These areas exhibit occasional inaccuracies or partial segmentation errors, reflecting a common challenge in both medical image segmentation and generative modelling, where complex and low-contrast anatomical boundaries are difficult to represent accurately. While these localized errors suggest that further improvements may be required to better capture fine-grained anatomical boundaries, pointing to potential directions for future work, it may also be a product of error propagation from the use of the segmentation DL model.

    \begin{figure}[tb]
            \centering
            \includegraphics[width=1.0\linewidth]{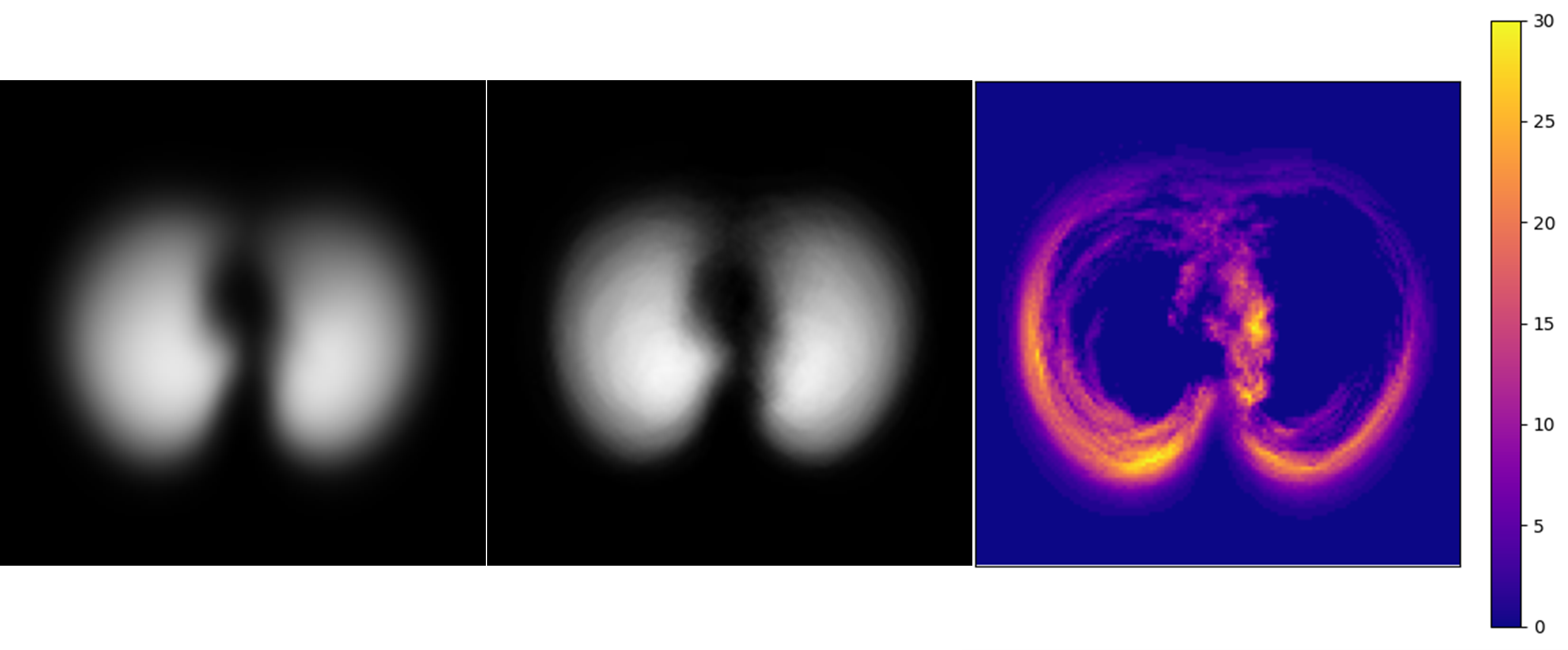}
            \caption{Visual representation of the computed intensity heatmaps for both the full LIDC-IDRI dataset (left) and a population of 100 synthetic samples (center). The difference in sample number (1018 vs. 100) explains the texture difference between the real and fake heatmaps (blurry vs blocky). The rightmost image displays the difference between the two heatmaps and a color scale, revealing the maximum discrepancy between intensity distributions at any given pixel is valued at 30 out of 255.}
            \label{fig:heatmap-results}
        \end{figure}

    The analysis of the heatmaps presented in figure~\ref{fig:heatmap-results} is crucial to understand the unconditional nature of the proposed generation method. Specifically, the discrepancy peaks being located at the lower lung region and near the hear indicate how having no control over the height value $Z$ of the produced sample can affect overall results. In other words, since the real sample heatmap on the left includes all the slices for $Z \in [30, 90]$ and there is no insight to be had into the $Z$ values of the fake slices, we can assume such discrepancies stem from structural size differences across different height values. In the real dataset, this value will make the real lung and heart either less (peripheral slices) or more prominent (middle slices), but as there is no control over this variable, produced samples aim for a range of different $Z$ values, which for the shown population seem to be more intermediate than peripheral.


    \subsection{Performance Across Generative Architectures}
    Beyond metric-specific comparisons, other consistent patterns emerge when analysing the architectural design choices across the evaluated generative pipelines.

    Multi-channel approaches are almost identical copies of their respective baseline architectures, with only the input and/or output layers of the networks changing dimensions, which introduces a negligible amount of learnable parameters to the generative pipeline. However, the data size increments proportionally to the number of HU intervals being used. This may lead to expect a performance drop for all multi-channel approaches compared to their respective baselines, since the the model capacity stays identical while the size the data distribution to be learned increases in complexity. This phenomenon was confirmed for the WGAN-GP and Score-based DMs approaches, where the multi-channel variations underperformed against their baselines for most of the HU windows, including full-range, across the various metrics. In contrast, the multi-channel VQVAE approaches consistently displayed better performance compared to its baseline across FID, MMD and Recall metrics and HU intervals. The VQVAE approaches likely benefit from the multi-channel formulation due to the discrete nature of their latent representation and the presence of a learned codebook. By quantizing latent features into a finite set of embeddings, the VQVAE effectively partitions the representation space into semantically meaningful regions, where, in this setting, different HU channels can be mapped to complementary subsets of the codebook, enabling specialization without enforcing direct competition within a shared continuous latent space.

    Across the evaluated metrics, the differences between multi-channel, multi-decoder, and multi-head VQVAE methods can be understood primarily as trade-offs between HU-specific specialization and global representational coherence. Despite forcing the encoding of the spatial coherence and elements of the various HU windows into a joint representation latent space, the multi-channel methodology also introduces challenges at the decoder level that can adversely affect both training stability and reconstruction quality. In particular, while spatial alignment across channels is generally well preserved, channel-specific textural details may be over-smoothed as gradients associated with disparate texture statistics interact through the shared decoder weights. In contrast, the multi-decoder approach targeted these limitations by decoupling the reconstruction of individual channels through multiple decoder branches, thereby reducing gradient interference and enabling channel-specific modelling and specialization in reconstructing the characteristics of corresponding HU intervals while preserving the shared spatial structure encoded by the latent representation. For the full-range CT images, the multi-decoder approaches achieved overall superior performance across all metrics compared to the multi-channel counterparts across the evaluated metrics. However, when performance is examined within individual HU domains, the relative ranking of the two approaches becomes considerably less stable. Depending on the specific HU interval and evaluation metric considered, either the multi-decoder or the multi-channel model may exhibit better performance, with no approach consistently dominating across all HU ranges. These observations may be attributed to the composite training loss in Equation \ref{eq:final-loss}, as the inclusion of the post-reconstruction term $\mathcal{L}_{\text{Post-Rec}}$ can bias optimization toward representations that are most effective for the employed reconstruction model, rather than uniformly optimizing fidelity within each individual HU domain that is imposed by the pre-reconstruction term $\mathcal{L}_{\text{Pre-Rec}}$. In this context, the multi-decoder networks provide greater flexibility by allowing channel-specific reconstructions to adapt independently, thereby reducing interference and enabling convergence to representations that better support the full-range mapping. Conversely, the multi-channel approaches enforce tighter coupling at the decoder level, which may promote more uniform behaviour across HU domains but can restrict the model’s ability to allocate representational capacity in a manner that is better aligned with the reconstruction model expected inputs for higher fidelity full-range CT image synthesis.

    The multi-head formulation represents a structured formulation between the fully shared multi-channel design and the fully separated multi-decoder approach. By introducing channel-specific encoder heads, the model can capture low-level, HU-dependent appearance characteristics before projecting them into a shared latent space that encodes the common spatial and semantic structure across HU domains. On the decoding side, this shared representation is processed by a common decoder backbone to enforce global coherence, after which channel-specific decoder heads reproduce HU-specific textural details in the final reconstructions. Empirically, this balance translated into the most consistent and reliable performance across the experiments, with the multi-head models achieving the best results for many subsets of metrics and HU windows, most notably in FID and MMD for full-range CT images, while remaining competitive in cases where it was not the top-performing method. This consistency suggests that the multi-head method strikes a favourable balance between joint representation learning and HU-specific specialization, mitigating the limitations observed in the multi-channel and multi-decoder architectures inspected earlier.

    \subsection{Model Complexity and Performance Trade-off}
    In addition to absolute reconstruction quality, model complexity plays a central role in determining the practical viability of generative architectures. Therefore, the relationship between performance and parameter count is examined across the evaluated models.

    The reconstruction models demonstrate a clear performance-complexity relationship, were increased capacity generally leads to improved reconstruction quality. While the simplest MLP-based models achieved great performance with extremely low parameter counts, they consistently underperform to their equivalently deep convolutional alternatives, particularly in distributional metrics such as FID and MMD. These substantial gains further highlight the advantage provided by the spatial inductive bias from the CNN reconstruction models compared to the fully-connected reconstruction networks. Moreover, it is worth noting that, although deeper convolutional configurations increase parameter count, the results obtained from these models translate to near-perfect HU-to-CT translations of the images at an extremely low memory footprint and parameter count.  

    Next, despite the WGAN-GP methods presenting the most lightweight networks, with generators and critics adding up to 6.8 million parameters, their generative fidelity remains substantially lower than that of the remaining methods. Similarly, the score-based diffusion models, with approximately 72.3 million parameters, represent the most parameter-heavy approach among the experimented, yet consistently underperform all VQVAE variants across all evaluation metrics. Together, these suggest that, for the multiple HU-window generative task, neither adversarial training nor diffusion-based sampling effectively translated increased model capacity into superior reconstruction quality or distributional alignment.

    Regarding VQVAE approaches, the multi-channel VQVAEs maintains identical parameter counts to its the baseline VQVAE at approximately 48.3 million but shows limited gains in generative performance, indicating that naively extending the input dimensionality does not substantially improve representational capacity and reconstruction. In contrast, the multi-decoder VQVAE significantly increases model size to 68.9 million parameters due to the replication of decoder components in the network. While this configuration yields improvements over simpler variants, the gains are not proportional to the additional complexity and memory usage, resulting in a less efficient approach. The multi-head VQVAE achieves a notably better balance between capacity and performance, where, with only a marginal increase from 48.3 million to 49.3 million parameters compared to its baseline, it consistently matches or outperforms both multi-channel and multi-decoder designs. By combining channel-specific encoder and decoder heads with a shared latent space, the multi-head VQVAE approach supports the creation of stronger joint latent representation of all HU-views and allows for the specialization of each decoding unit into HU-specific restoration while avoiding the parameter count overhead of fully independent decoders. Consequently, it achieves the most favourable performance-complexity trade-off across the evaluated approaches.

    \subsection{Decoupled Generation and Reconstruction}
    An additional practical implication of the proposed pipeline lies in the explicit decoupling of the generative process for HU-windowed views from full-range CT reconstruction and downstream usage.
    
    In this formulation, the HU samples act as intermediate representation that is generated once, store persistently, and subsequently reconstructed by a reconstruction model to recover a full-range representation on demand. This design is motivated by the pronounced asymmetry between the generative and reconstruction tasks computational costs, since the generative models are computationally intensive, both in terms of training and inference, whereas the reconstruction networks are compact and efficient. By isolating the expensive synthesis step, large collections of HU views can be produced offline and archived, while reconstruction can be performed dynamically as required by a given task, visualization protocol, or any other specialized application.

    In contrast, the finite number of HU-clipped samples that can be generated and stored imposes concerns regarding potential limitation of the diversity of full-range outputs. However, it rests on the assumption that the primary objective is to maximize the amount of full-range CT images, ignoring the utility of the HU-filtered images produced by the generative step of the proposed pipeline.


\section{Conclusion}
\label{sec:conclusion}
In essence, this work introduced a novel framework for full-range lung CT scan slices synthesis based on modelling disjunctive HU domains, establishing HU-decomposition generation as a viable paradigm for CT image synthesis. By reformulating CT generation as a the synthesis of multiple HU-windowed representations followed by a learned reconstruction step, the proposed methodology addresses the complexity of modelling the full CT intensity distribution.

Initial results demonstrate that full-range CT images can be accurately reconstructed from a limited set of non-overlapping HU intervals using extremely lightweight neural networks, achieving excellent reconstruction accuracy, distributional fidelity and structural similarity results with negligible computation cost. After extensive experimental evaluation among the explored generative approaches, the VQVAE-based methods consistently outperformed their baseline, contrary to the WGAN-GP and score-based DM ones. In particular, the multi-head VQVAE architecture achieves the most favourable balance between performance and complexity, delivering superior FID and MMD scores for full-range reconstructions while maintaining competitive Precision and Recall across all HU intervals, as well as adequate structural variety within the generated samples.

Beyond quantitative gains, the qualitative segmentation task assessment further supports the fidelity and utility of the generated images, showing that synthetic CT scans produced by the proposed pipeline are compatible with models trained on real data, although  some localized imperfections remain in anatomically complex regions, which point towards future improvements in fine-grained boundary modelling. Furthermore, the proposed framework offers the ability to disconnected the computationally expensive generative process of the pipeline from the lightweight full-range reconstruction process, enabling the HU-windowed samples to become reusable for full-range CT slices reconstruction on demand.

Future work may extend this framework to three-dimensional medical scan volumes, additional HU decompositions, or other imaging modalities where intensity-based semantics are meaningful, further reinforcing the role of clinically informed design in generative medical imaging research.


\section*{Acknowledgments}
This work is supported by the European Commission funded PHASE IV AI project (Grant Agreement No. 101095384) under the European Union's Horizon Europe research and innovation programme.

Moreover, the authors acknowledge the National Cancer Institute and the Foundation for the National Institutes of Health, and their critical role in the creation of the free publicly available LIDC/IDRI Database used in this study.

\bibliographystyle{unsrt}
\bibliography{references}  

\end{document}